\documentclass[10pt,twocolumn,letterpaper]{article}

\usepackage[applications]{wacv}      

\usepackage[accsupp]{axessibility}  

\usepackage{amsmath}
\usepackage{amssymb}
\usepackage{amsthm}
\usepackage{bm}

\usepackage{wrapfig}
\usepackage{booktabs}
\usepackage{multirow}
\usepackage{makecell}
\usepackage{caption}

\usepackage{graphicx}
\usepackage[table]{xcolor}

\usepackage{subcaption}

\usepackage{bbding}

\usepackage{algorithm}
\usepackage{algorithmic}  

\newcommand{\argmin}{\mathop{\mathrm{argmin}}}

\newcommand{\bench}{M-ErasureBench}
\newcommand{\highlight}[1]{\cellcolor{orange!18}{#1}}
\newcommand{\highlightavg}[1]{\cellcolor{red!18}{#1}}
\newcommand{\improve}[1]{\textcolor{green!50!Black}{#1}}
\newcommand{\worse}[1]{\textcolor{red}{#1}}
\newcommand{\highlighttext}[1]{\colorbox{orange!18}{#1}}
\newcommand{\highlightavgtext}[1]{\colorbox{red!18}{#1}}

\definecolor{darkgreen}{rgb}{0.0,0.5,0.0}
\definecolor{crimson}{rgb}{0.86, 0.08, 0.24}
\definecolor{royalblue}{rgb}{0.25, 0.35, 0.74}
\definecolor{DarkCyan}{rgb}{0.0, 0.54, 0.54}
\definecolor{Gray}{gray}{0.95}
\definecolor{ChromeYellow}{rgb}{1.0, 0.65, 0.0}

\definecolor{wacvblue}{rgb}{0.21,0.49,0.74}
\usepackage[pagebackref,breaklinks,colorlinks,allcolors=wacvblue]{hyperref}

\def\wacvPaperID{2162} 
\def\confName{WACV}
\def\confYear{2026}

\title{\bench: A Comprehensive Multimodal Evaluation Benchmark for Concept Erasure in Diffusion Models}

\author{
Ju-Hsuan Weng$^{1, 2*}$
\quad
Jia-Wei Liao$^{1, 2}$\thanks{denotes equal contribution.}
\quad
Cheng-Fu Chou$^{1}$
\qquad
Jun-Cheng Chen$^{2}$
\vspace{0.5em} \\
$^{1}$ National Taiwan University \\
$^{2}$ Research Center for Information Technology Innovation, Academia Sinica
}

\begin{document}

\maketitle
\begin{abstract}
Text-to-image diffusion models may generate harmful or copyrighted content, motivating research on concept erasure. However, existing approaches primarily focus on erasing concepts from text prompts, overlooking other input modalities that are increasingly critical in real-world applications such as image editing and personalized generation. These modalities can become attack surfaces, where erased concepts re-emerge despite defenses. To bridge this gap, we introduce \bench{}, a novel multimodal evaluation framework that systematically benchmarks concept erasure methods across three input modalities: text prompts, learned embeddings, and inverted latents. For the latter two, we evaluate both white-box and black-box access, yielding five evaluation scenarios. Our analysis shows that existing methods achieve strong erasure performance against text prompts but largely fail under learned embeddings and inverted latents, with Concept Reproduction Rate (CRR) exceeding 90\% in the white-box setting. To address these vulnerabilities, we propose IRECE (Inference-time Robustness Enhancement for Concept Erasure), a plug-and-play module that localizes target concepts via cross-attention and perturbs the associated latents during denoising. Experiments demonstrate that IRECE consistently restores robustness, reducing CRR by up to 40\% under the most challenging white-box latent inversion scenario, while preserving visual quality. To the best of our knowledge, \bench{} provides the first comprehensive benchmark of concept erasure beyond text prompts. Together with IRECE, our benchmark offers practical safeguards for building more reliable protective generative models.
\end{abstract}
    
\section{Introduction}

The contemporary diffusion models~\cite{sohl2015pdm,ho2020ddpm,rombach2022ldm,esser2024sd3,ramesh2022dalle2,nichol2022glide,saharia2022imagen} have demonstrated remarkable progress in high-quality and versatile content generation, supporting tasks such as image synthesis~\cite{dhariwal2021cg,ho2021cfg,rombach2022ldm,zhang2023controlnet}, image editing~\cite{meng2022sdedit,hertz2022p2p,brooks2023instructpix2pix,hsiao2025tfti2ti}, personalized generation~\cite{gal2023textualinv,ruiz2023dreambooth}, and style transfer~\cite{chung2024styletransfer_1,zhang2023styletransfer_2}. However, training on large-scale uncurated datasets makes them prone to reproducing copyrighted~\cite{blaszczyk2024copyright_1,grynbaum2023copyright_2} or inappropriate content~\cite{hunter2023nsfw_1}. Retraining on filtered datasets offers a direct solution, but it is costly and often degrades generative quality~\cite{esser2024sd3}. Recent studies therefore explore concept erasure~\cite{fan2024salun,zhang2024fmn,gandikota2023esd,gandikota2024uce,kumari2023ac,huang2024receler,lu2024mace,basu2023diffquickfix,concept2025cpe,thakral2025fade,wu2025erasediff,wang2025adavd}, which aims to prevent text-to-image diffusion models from generating harmful or copyrighted content
by suppressing specific concepts through fine-tuning cross-attention layers of diffusion models
without retraining from scratch.

Despite their effectiveness under text prompts, existing methods reveal critical weaknesses. In practice, users often rely on learned embeddings from personalization techniques~\cite{gal2023textualinv,ruiz2023dreambooth} or noisy latent from inversion methods~\cite{dhariwal2021cg, mokady2023nulltextinv}, which fall outside the assumptions of text-based concept erasure. Our analysis shows that while suppression is reliable for basic prompts, concepts frequently re-emerge with learned embeddings or inverted latents, reaching Concept Reproduction
Rate (CRR) over 90\% in white-box setting with unconditional prompt. 

Geroge \emph{et al.}~\cite{george2025illusion} also found that the erased concepts can be revived through fine-tuning the model with a few samples. This indicates that current methods primarily disrupt text–image alignment rather than fully removing concepts.  Moreover, although adversarial prompt attacks~\cite{chin2024p4d,pham2024cce} expose vulnerabilities, defenses based on adversarial training~\cite{huang2024receler} remain limited to textual inputs, underscoring the need to explore robustness beyond the text space. This raises our central research question: \emph{How robust are concept erasure methods across different input modalities, and can their vulnerabilities be mitigated without retraining?}

Motivated by these limitations, we introduce \bench{}, a novel multimodal evaluation benchmark that systematically evaluates the robustness of the concept erasure methods across three representative input settings for text-to-image diffusion models:
text prompts, learned embeddings, and inverted latents, under both white-box and black-box settings, respectively. With the extensive evaluations from \bench{}, the results demonstrate that the state-of-the-art concept erasure methods are still vulnerable to various inference-time attack methods. Building upon these insights, we further propose Inference-time Robustness Enhancement for Concept Erasure (IRECE), a plug-and-play module that localizes target concepts via cross-attention and selectively perturbs associated latents during denoising without retraining.

Our contributions are summarized as follows:
\begin{itemize}
    \item We introduce \bench{}, the first comprehensive multimodal evaluation benchmark for concept erasure of text-to-image diffusion models, covering three input modalities: text prompts, learned embeddings, and inverted latents, under both white- and black-box settings.
    \item Our study reveals that although existing methods are effective for text prompts, they largely fail under learned embeddings and latent inversion, with CRR exceeding 90\% in the white-box setting with the unconditional prompt when adversarial training is absent. This indicates that current approaches disrupt alignment rather than fully erasing concepts.
    \item We propose IRECE, a plug-and-play module that requires no retraining. By localizing target concepts via cross-attention and perturbing associated latents, IRECE reduces CRR by up to 40\% under the white-box setting with the unconditional prompt, while maintaining visual quality. This provides a practical enhancement for safer generative models.
\end{itemize}

\begin{figure*}
    \centering
    \includegraphics[width=0.9\linewidth]{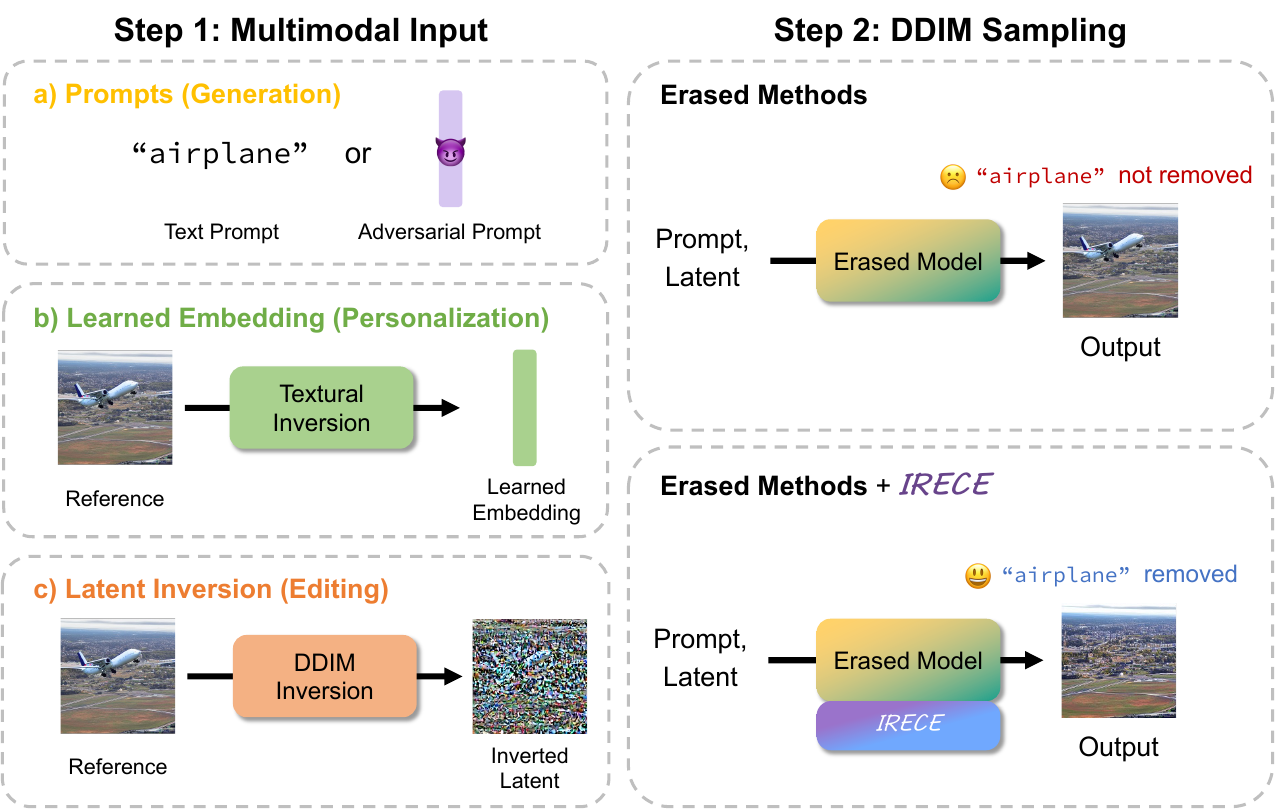}
    \caption{
    \textbf{Illustration of ErasureBench and IRECE.} We evaluate concept-erasure methods on text prompts, learned embeddings, and inverted latents. While erased models suppress concepts under simple text prompts, they fail in the other two settings, where the target concept (``airplane") re-emerges. Our proposed IRECE module, applied at inference time, restores robustness and removes the target concept without retraining.
    \vspace{-10pt}
    }
    \label{fig:teaser}
\end{figure*}

\section{Related Works}

\subsection{Concept Erasure in Diffusion Models}
Text-to-image diffusion models~\cite{rombach2022ldm,ramesh2022dalle2,nichol2022glide,saharia2022imagen} have achieved impressive performance in high-quality image generation, but their reliance on large-scale web data introduces risks of reproducing copyrighted or harmful content~\cite{blaszczyk2024copyright_1,grynbaum2023copyright_2,hunter2023nsfw_1}. Retraining on curated datasets can mitigate these risks, yet it is costly and often compromises generation quality~\cite{esser2024sd3}, motivating research on concept erasure. Most existing approaches~\cite{fan2024salun,zhang2024fmn,gandikota2023esd,gandikota2024uce,kumari2023ac,huang2024receler,lu2024mace,basu2023diffquickfix} fine-tune the UNet~\cite{ronneberger2015unet}, particularly its cross-attention layers~\cite{vaswani2017transformer}, and can be broadly categorized into anchor-based and anchor-free methods. Anchor-based approaches remove concepts by aligning their embeddings with substitutes (e.g., mapping ``weapon" to ``tool"), while anchor-free approaches steer generation away from the target concept without replacement. Representative methods include ESD~\cite{gandikota2023esd}, which applies negative guidance to suppress target concepts; UCE~\cite{gandikota2024uce}, which edits cross-attention projections in closed form to remove specific alignments; and Receler~\cite{huang2024receler}, which enhances robustness via adversarial training and parameter-efficient fine-tuning~\cite{hu2022lora}, providing tunable control over erasure strength. While effective for textual prompts, these text-driven methods struggle to generalize to non-textual inputs such as learned embeddings or inverted latents. EraseBench~\cite{amara2025erasebench} provides a valuable benchmark for assessing post-erasure side effects within text prompts, but it does not evaluate robustness across different input modalities. This motivates our proposed framework, which systematically examines the performance of concept erasure methods under diverse modality conditions.

\subsection{Adversarial Attacks on Concept Erasure}
Recent studies~\cite{truong2025qfattack,zhang2024unlearndiffatk,tsai2024ring-a-bell,pham2024cce,tsai2024ring-a-bell,yang2024mma} reveal that concept erasure methods remain vulnerable to adversarial attacks, most commonly through crafted prompts that regenerate erased concepts. For example, P4D~\cite{chin2024p4d} proposes a white-box attack by aligning noise predictions with those of the original model, while Receler~\cite{huang2024receler} incorporates adversarial prompt optimization to improve resilience. Ring-A-Bell~\cite{tsai2024ring-a-bell} constructs adversarial prompts using concept vectors derived from contrasts between positive and negative prompts. Beyond purely textual attacks, Concept Inversion~\cite{pham2024cce} leverages Textual Inversion~\cite{gal2023textualinv} to learn embeddings from reference images and probe whether erased concepts reappear. Despite these advances, most existing attacks remain text-driven, and their effectiveness diminishes when models adopt adversarial training, underscoring the need to explore robustness against non-textual pathways.

\section{Preliminaries}

\subsection{Denoising Diffusion Implicit Models (DDIM)}
Diffusion models~\cite{sohl2015pdm,ho2020ddpm, song2020score} synthesize images by progressively denoising a latent from random noise. Denoising Diffusion Implicit Models (DDIM)~\cite{song2021ddim} extend this process with a deterministic formulation, enabling faster sampling without compromising quality. Given a latent $\bm{x}_t$ at timestep $t$, the update to the previous latent $\bm{x}_{t-1}$ is expressed as
\begin{align*}
    \bm{x}_{t-1}
    &= \operatorname{DDIMStep}(\bm{x}_t, \bm{c}, t, \theta) \\
    &:= \sqrt{\frac{\bar{\alpha}_{t-1}}{\bar{\alpha}_t}}
    \left( \bm{x}_t - \sqrt{1 - \bar{\alpha}_t} \bm{\epsilon}_{\theta}(\bm{x}_t, t, \bm{c}) \right) \\
    & +
    \sqrt{1 - \bar{\alpha}_{t-1}} \bm{\epsilon}_{\theta}(\bm{x}_t, t, \bm{c}),
\end{align*}
where $\bar{\alpha}_t$ is the cumulative noise schedule, $\bm{\epsilon}_\theta$ is the denoiser, and $\bm{c}$ is the conditioning prompt. The full denoising trajectory from $T$ to $0$ is compactly written as
\begin{equation*}
\bm{x}_{0} = \operatorname{DDIMProcess}(\bm{x}_T, \bm{c}, T, \theta).
\end{equation*}
Although diffusion models excel at generating novel content, many applications require editing existing images. Early work such as SDEdit~\cite{meng2022sdedit} showed that applying the forward SDE for editing often distorts image content, since injected noise may overwrite details or alter semantic objects. This motivated the development of inversion techniques that map real images back into the latent space of diffusion models. Among them, DDIM inversion~\cite{dhariwal2021cg} offers a principled mechanism: by reversing the denoising process, it reconstructs the noisy latent trajectory of a reference image, which can then be reused for resampling or manipulation. Formally, given a latent $\bm{x}_t$ at timestep $t$, the inversion update is expressed as
\begin{align*}
\bm{x}_{t+1}
&= \operatorname{DDIMInvStep}(\bm{x}_t, \bm{c}, t, \theta) \\
&:= \sqrt{\frac{\bar{\alpha}_{t+1}}{ \bar{\alpha}_t }} \left( \bm{x}_t - \sqrt{1 - \bar{\alpha}_t} \bm{\epsilon}_\theta(\bm{x}_t, t, \bm{c}) \right) \\
&+ \sqrt{1 - \bar{\alpha}_{t+1}} \bm{\epsilon}_\theta(\bm{x}_t, t, \bm{c}),
\end{align*}
and the full inversion trajectory is denoted as
\begin{equation*}
\bm{x}_{T} = \operatorname{DDIMInvProcess}(\bm{x}_0, \bm{c}, T, \theta).
\end{equation*}

\subsection{Textual Inversion}
While natural language provides a flexible interface for diffusion models, its ambiguity often limits precise control. Personalization techniques address this by enabling users to inject concepts from a few reference images. Among them, Textual Inversion (TI)~\cite{gal2023textualinv} stands out as a lightweight approach that avoids retraining the diffusion model. From a reference image $\bm{x}_0$, TI learns a dedicated embedding $\bm{e}^*$ associated with a placeholder token. Once trained, prompts containing this token reliably evoke the target concept. Formally, $\bm{e}^*$ is optimized to minimize the reconstruction error in the frozen diffusion model:
\begin{align*}
    \bm{e}^*
    &= \operatorname{TI}(\bm{x}_0, \bm{c}, \theta) \\
    &:=
    \argmin_{\bm{e}} \mathbb{E}_{t, \bm{\epsilon}_t, \bm{x}_t|\bm{x}_0} \left[ \|\bm{\epsilon}_t - \bm{\epsilon}_\theta(\bm{x}_t, t, [\bm{c}, \bm{e}]) \|_2^2 \right],
\end{align*}
where $\bm{x}_t$ is the noisy latent at timestep $t$, $\bm{\epsilon}_t \sim \mathcal{N}(\bm{0}, \bm{I})$ is Gaussian noise at timestep $t$, $\bm{\epsilon}_\theta$ is the frozen denoiser, and $[\bm{c}, \bm{e}]$ denotes the conditioning vector obtained by inserting the learned embedding $\bm{e}$ into the prompt $\bm{c}$.

\section{Multimodal Concept Erasure Evaluation} \label{sec:mm_eval_framework}

Most concept erasure methods are evaluated only on text prompts. However, in real-world scenarios users may interact with diffusion models through richer modalities such as learned embeddings or image-based latents. To capture these cases, we design a comprehensive multimodal evaluation framework (Figure~\ref{fig:teaser}) with three settings: text prompts, learned embeddings, and latent inversion.

\subsection{Text Prompt Evaluation}
In typical usage, users interact with diffusion models through natural language inputs~\cite{rombach2022ldm}. We therefore begin by evaluating erased models with basic prompts explicitly containing the target concept. To further assess robustness, we incorporate adversarial prompts, as prior work~\cite{chin2024p4d,tsai2024ring-a-bell} has shown that erased models remain vulnerable to prompt-based attacks that can bypass suppression mechanisms. In our setup, adversarial prompts generated by Ring-A-Bell~\cite{tsai2024ring-a-bell} serve as a baseline for comparison.

\subsection{Learned Embedding Evaluation}
While standard text-to-image diffusion models such as Stable Diffusion~\cite{rombach2022ldm} support text prompts, they are limited in personalizing generation for specific objects. To overcome this limitation, prior work has explored learning dedicated embeddings that can be combined with prompts for user-specific customization. To evaluate robustness under this setting, we adopt Textual Inversion (TI)~\cite{gal2023textualinv}, which learns an embedding $\bm{e}^*$ from reference images and conditions generation by augmenting the prompt embedding $\bm{c}$ with this learned embedding. However, in realistic scenarios, access to the parameters of the erased model may be restricted. We therefore consider three variants:

\paragraph{White-box.}
TI is trained directly on the erased model $\theta_\text{era}$, and the learned embedding is then used for generation.
\begin{align*}
    \bm{e}^* &= \operatorname{TI}(\bm{x}_0, \bm{c}, \theta_\text{era}), \\
    \bm{x}_T &\sim \mathcal{N}(\bm{0}, \bm{I}), \\
    \bm{x}_{0}^* &= \operatorname{DDIMProcess}(\bm{x}_T, [\bm{c}, \bm{e}^*], T, \theta_\text{era}).
\end{align*}

\paragraph{(Surrogate-based) Black-box.}
TI is trained on a standard model $\theta_\text{std}$ as the surrogate, 
sharing the same backbone architecture as the erased model $\theta_\text{era}$, to approximate its behavior when direct access is unavailable. The learned embedding is then transferred to $\theta_\text{era}$ for generation.

\begin{align*}
    \bm{e}^* &= \operatorname{TI}(\bm{x}_0, \bm{c}, \theta_\text{std}), \\
    \bm{x}_T &\sim \mathcal{N}(\bm{0}, \bm{I}), \\
    \bm{x}_{0}^* &= \operatorname{DDIMProcess}(\bm{x}_T, [\bm{c}, \bm{e}^*], T, \theta_\text{era}).
\end{align*}

\paragraph{(Surrogate-based) Black-box with Perturbations.}
To further probe robustness, random noise $\bm{\tau} \sim \mathcal{U}([-a, a]^d)$ is added to the reference image before TI training to induce semantic shifts in $\bm{e}^*$, where $a>0$ and $d={\operatorname{dim}(\bm{x}_0)}$.
\begin{align*}
    \Tilde{\bm{x}}_0 &= \operatorname{clip}(\bm{x}_0 + \bm{\tau}, -1, 1), \\
    \bm{e}^* &= \operatorname{TI}(\Tilde{\bm{x}}_0, \bm{c}, \theta_\text{std}), \\
    \bm{x}_T &\sim \mathcal{N}(\bm{0}, \bm{I}), \\
    \bm{x}_{0}^* &= \operatorname{DDIMProcess}(\bm{x}_T, [\bm{c}, \bm{e}^*], T, \theta_\text{era}).
\end{align*}

\subsection{Latent Inversion Evaluation}
Latent inversion is widely used in image editing pipelines, where techniques such as DDIM inversion~\cite{dhariwal2021cg} recover the latent representation of a reference image prior to resampling. Evaluating concept erasure under this setting is critical because the inverted latent inherently preserves concept-related information. In our framework, we apply DDIM inversion to map a reference image to an initial noisy latent and then combine it with a sampling prompt during generation, where $\bm{c}_\text{inv}$ denotes the prompt used for inversion and $\bm{c}_\text{sam}$ denotes the prompt used for resampling. This setup allows us to test whether erased concepts re-emerge when generation is initialized from concept-containing latents. We consider two configurations:

\paragraph{White-box.}
Both inversion and resampling are performed on the erased model $\theta_\text{era}$.
\begin{align*}
    \bm{x}_{T} &= \operatorname{DDIMInvProcess}(\bm{x}_0, \bm{c}_\text{inv}, T, \theta_\text{era}), \\
    \bm{x}_{0}^* &= \operatorname{DDIMProcess}(\bm{x}_T, \bm{c}_\text{sam}, T, \theta_\text{era}).
\end{align*}

\paragraph{(Surrogate-based) Black-box.}
Inversion is performed on a surrogate model $\theta_\text{std}$ with the same backbone architecture, while resampling is executed on the erased model $\theta_\text{era}$.
\begin{align*}
    \bm{x}_{T} &= \operatorname{DDIMInvProcess}(\bm{x}_0, \bm{c}_\text{inv}, T, \theta_\text{std}), \\
    \bm{x}_{0}^* &= \operatorname{DDIMProcess}(\bm{x}_T, \bm{c}_\text{sam}, T, \theta_\text{era}).
\end{align*}

For evaluation, we adopt the four prompt-pair strategies in Table 1 of the Appendix, covering unconditional, generic, coarse, and explicit target cases.

\begin{figure*}[t]
    \centering
    \begin{subfigure}[t]{0.45\textwidth}
        \centering
        \includegraphics[width=\linewidth]{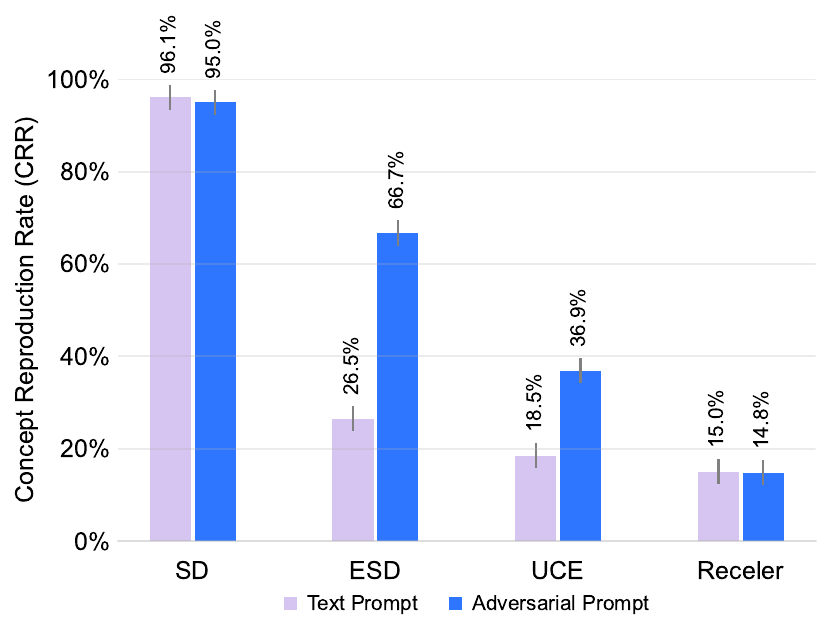}
        \caption{Text Prompt}
        \label{fig:text_prompt_eval}
    \end{subfigure}
    \hfill
    \begin{subfigure}[t]{0.45\textwidth}
        \centering
        \includegraphics[width=\linewidth]{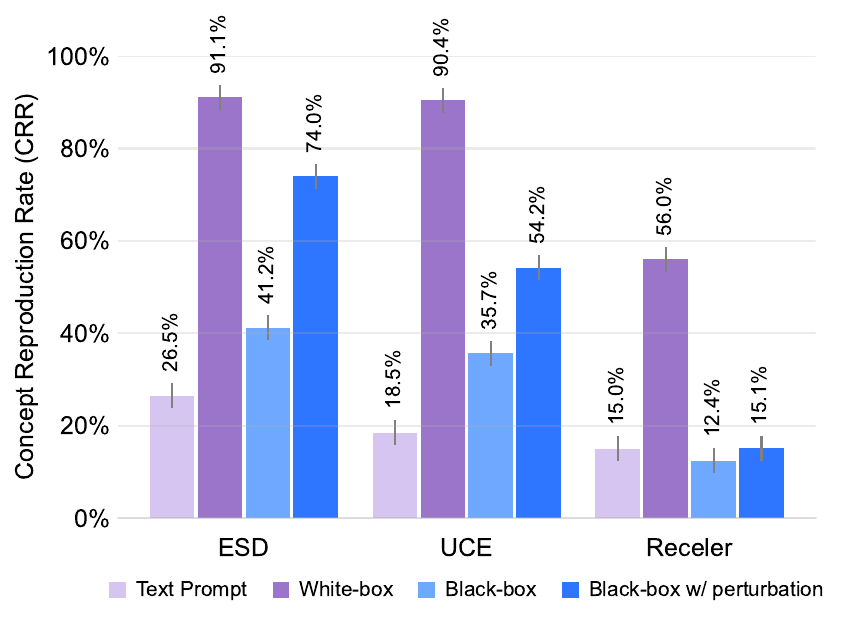}
        \caption{Learned Embedding}
        \label{fig:learned_embedding_eval}
    \end{subfigure}
    \caption{
    \textbf{Concept Reproduction Rate (CRR) for concept-erasure methods under two evaluation settings: (a) text prompts and (b) learned embeddings.} In (a), bar colors distinguish between original text prompts and adversarial prompts. In (b), bar colors denote white-box, black-box, and black-box with perturbation settings, with results from original text prompts included as a reference.
    }
    \label{fig:text_prompt_and_learned_emb_eval}
    \vspace{-5pt}
\end{figure*}

\begin{figure}
    \centering
    \includegraphics[width=\linewidth]{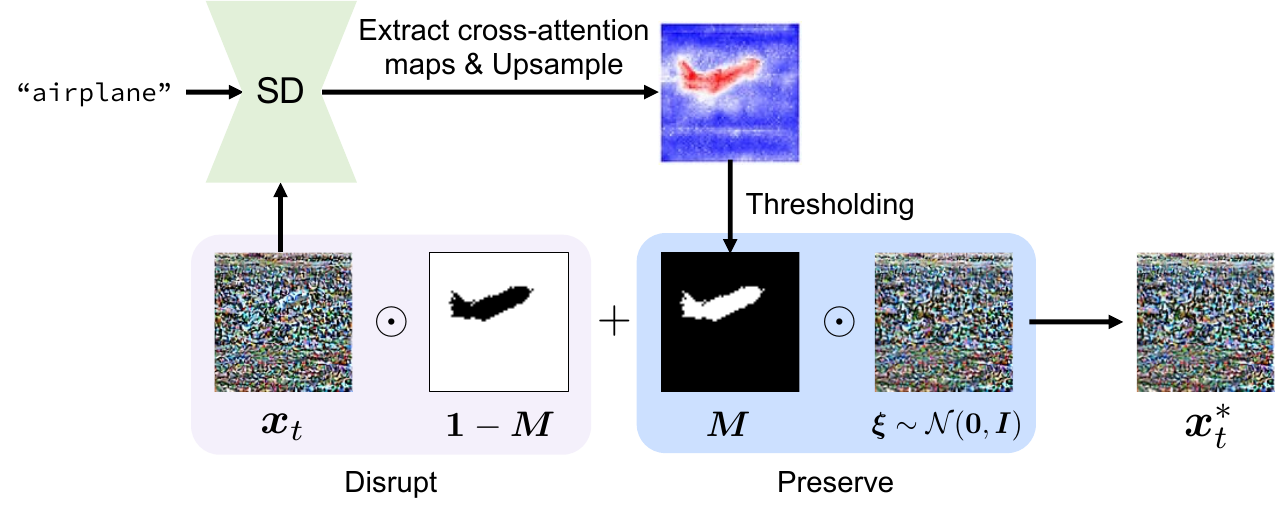}
    \caption{
    \textbf{Overview of IRECE.} Given a latent $\bm{x}_t$, SD provides cross-attention maps that identify regions corresponding to the target concept. After thresholding the aggregated map to obtain a binary mask $\bm{M}$, IRECE replaces the masked regions with Gaussian noise $\bm{\xi}$, forming an updated latent $\bm{x}_t^*$ that removes concept-related information while leaving surrounding content intact.
    \vspace{-5pt}
    }
    \label{fig:irece}
\end{figure}

\section{Inference-time Robustness Enhancement for Concept Erasure (IRECE)} \label{sec:irece}

To improve the reliability of erased models, we propose \emph{IRECE (Inference-time Robustness Enhancement for Concept Erasure)}, a plug-and-play module that strengthens robustness without retraining (Figure~\ref{fig:irece}). Inspired by prior work on concept localization through attention maps~\cite{lo2024distraction}, the key idea is to \emph{directly disrupt the latent regions responsible for encoding the erased concept} during inference, while leaving the rest of the image unaffected. The procedure begins with the initial noisy latent $\bm{x}_T$, which is progressively denoised by the erased model $\theta_{\text{era}}$ under the guidance of the sample prompt embedding $\bm{c}_{\text{sam}}$. At a chosen intervention step $t^*$, we detect spatial regions linked to the target concept embedding $\bm{c}_{\text{tgt}}$ by leveraging cross-attention maps $A_\text{cross}^\ell$ extracted from each layer $\ell$ of the standard model $\theta_{\text{std}}$. Since attention maps vary in resolution across layers, they are first upsampled to a common size and then aggregated:
\begin{equation*}
    \bm{A} = \sum_{\ell=1}^L \operatorname{Upsample}
    \left(A_\text{cross}^\ell(\bm{x}_t, \bm{c}_{\text{tgt}}; \theta_\text{std})\right),
\end{equation*}
From $\bm{A}$, a binary mask $\bm{M} \in \{0,1\}^{H \times W}$ is constructed by thresholding with parameter $\tau$, localizing the pixels most associated with the erased concept:
\begin{equation*}
\bm{M}(i,j) = \begin{cases}
1, & \bm{A}(i,j) \geq \tau, \\
0, & \text{otherwise}.
\end{cases}
\end{equation*}
Within these masked regions, the latent representation is perturbed by injecting Gaussian noise $\bm{\xi}_t \sim \mathcal{N}(\bm{0}, \bm{I})$, disrupting target-related information while largely preserving surrounding content:
\begin{equation*}
\bm{x}_t^* = (\bm{1}-\bm{M}) \odot \bm{x}_t + \bm{M} \odot \bm{\xi}_t,
\end{equation*}
Denoising resumes from $\bm{x}_t^*$ according to
\begin{equation*}
    \bm{x}_{t-1}
    = \begin{cases}
        \operatorname{DDIMStep}(\bm{x}_t^*, \bm{c}_\text{sam}, t, \theta_\text{era}), & \text{ if } t = t^*, \\
        \operatorname{DDIMStep}(\bm{x}_t, \bm{c}_\text{sam}, t, \theta_\text{era}), & \text{ otherwise}.
    \end{cases}
\end{equation*}
The process continues to $t=0$ to obtain the final image $\bm{x}_0^*$.

By localizing erased concepts through attention, masking their spatial footprint, and replacing the corrupted latent regions with noise, IRECE prevents erased concepts from re-emerging while maintaining visual coherence elsewhere. This plug-and-play design requires no retraining and operates entirely at inference time, and can naturally extend to tasks such as targeted object removal or replacement.

\section{Experiments}

\begin{figure*}[t]
    \centering
    \begin{subfigure}[t]{0.32\textwidth}
        \centering
        \includegraphics[width=\linewidth]{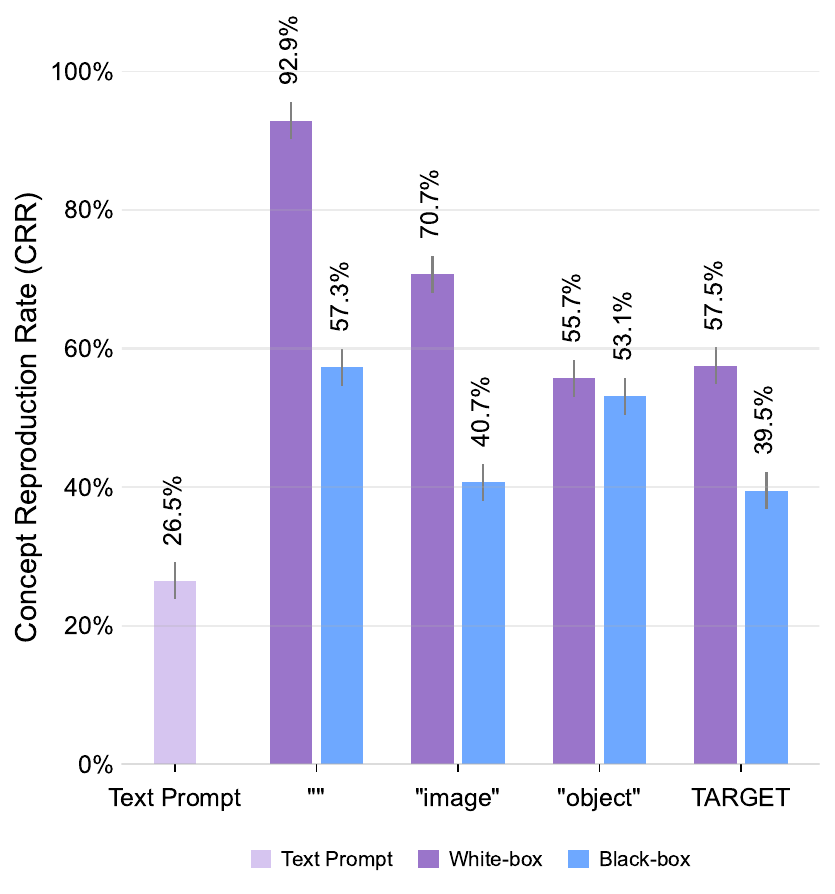}
        \caption{ESD}
        \label{fig:latent_esd_eval}
    \end{subfigure}
    \hfill
    \begin{subfigure}[t]{0.32\textwidth}
        \centering
        \includegraphics[width=\linewidth]{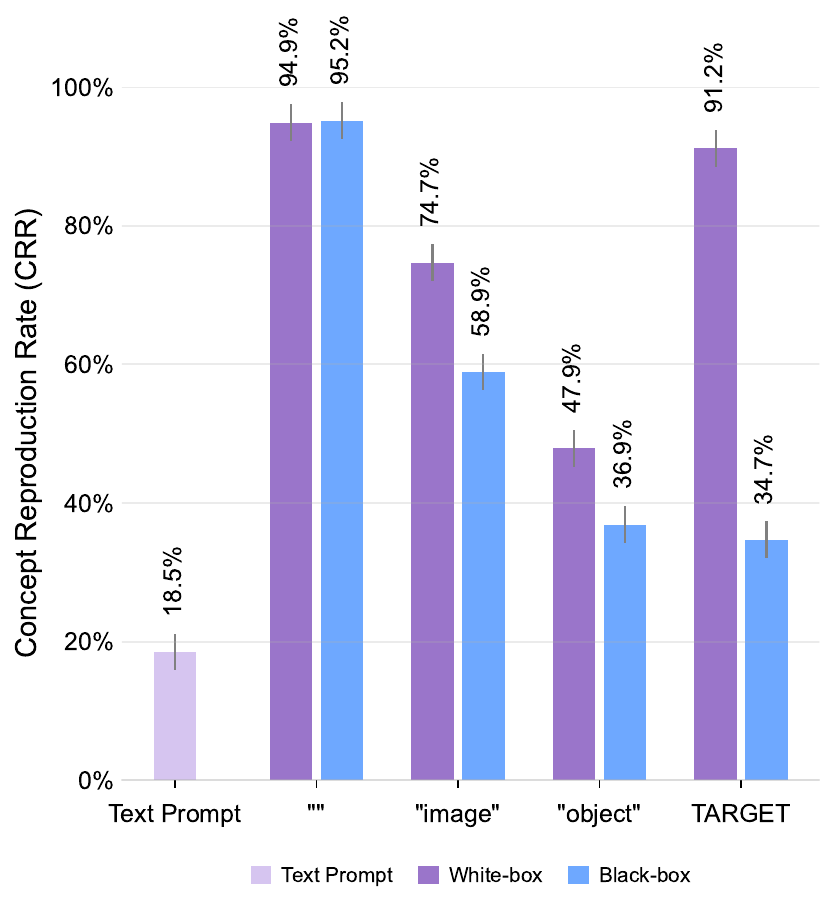}
        \caption{UCE}
        \label{fig:latent_uce_eval}
    \end{subfigure}
    \hfill
    \begin{subfigure}[t]{0.32\textwidth}
        \centering
        \includegraphics[width=\linewidth]{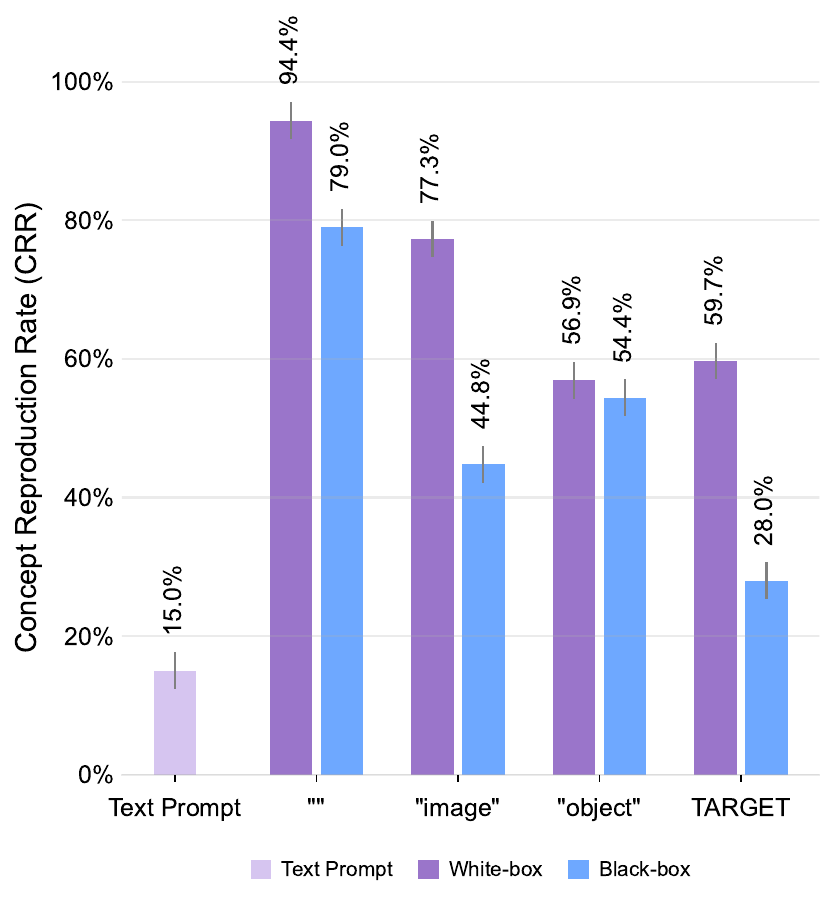}
        \caption{Receler}
        \label{fig:latent_receler_eval}
    \end{subfigure}
    \caption{
    \textbf{Concept Reproduction Rate (CRR) under latent-inversion evaluation.} Subfigures correspond to three representative concept-erasure methods: (a) ESD, (b) UCE, and (c) Receler. In each subfigure, horizontal axis groups denote different prompt types (``", ``image", ``object" and \texttt{TARGET}), and bar colors indicate white-box versus black-box evaluation settings. 
    }
    \label{fig:latent_eval}
    \vspace{-10pt}
\end{figure*}

\subsection{Experimental Setup}

\paragraph{Dataset.}
We adopt Stable Diffusion v1.4 (SD v1.4)~\cite{rombach2022ldm} as the base model and evaluate three representative erasure methods: ESD~\cite{gandikota2023esd}, UCE~\cite{gandikota2024uce}, and Receler~\cite{huang2024receler}. Following prior work, we use CIFAR-10~\cite{krizhevsky2009cifa10} class labels as target concepts. For evaluation, we construct four datasets.

\begin{itemize}
    \item \textbf{SD-Normal:} Constructed by generating images from SD using five prompt templates: ``An image of \texttt{TARGET}", ``A painting of \texttt{TARGET}", ``A picture of \texttt{TARGET}", ``A photo of \texttt{TARGET}", and simply ``\texttt{TARGET}", each instantiated with 30 random seeds, yielding 150 prompts per class.
    \item \textbf{SD-AdvPrompt:} Constructed by attacking the SD-Normal templates with Ring-A-Bell~\cite{tsai2024ring-a-bell} using the method's official configuration.
    \item \textbf{SD-TI:} Constructed by learning a special embedding for each reference image in the SD-Normal set via Textual Inversion~\cite{gal2023textualinv}, optimized with a learning rate of 5e-4 for 1500 optimization steps, and then inserting the learned token into the same prompt templates.
    \item \textbf{SD-LatentInv:} Constructed by applying DDIM inversion~\cite{dhariwal2021cg} to images in the SD-Normal dataset to obtain their initial latents.
\end{itemize}

\paragraph{Evaluation Metrics.}
We assess concept erasure using GroundingDINO~\cite{liu2024groundingdino} and report the \emph{Concept Reproduction Rate (CRR)}, defined as the percentage of generated samples containing the erased concept. Lower CRR indicates stronger suppression.

\subsection{Evaluation Results}

\subsubsection{Text Prompt Evaluation}
We evaluate concept erasure methods on both the SD-Normal and SD-AdvPrompt datasets, with results in Figure~\ref{fig:text_prompt_and_learned_emb_eval}(a). Under the standard text prompt setting, SD exhibits a high CRR of 96.1\%, indicating that without erasure the model consistently reproduces the target concept. In contrast, erasure methods markedly suppress reproduction: ESD reduces CRR to 26.5\% (-69.6\%), UCE to 18.5\% (-77.6\%), and Receler to 15.0\% (-81.1\%). These results demonstrate the effectiveness of current approaches in the text prompt evaluation. However, under adversarial prompts, SD still maintains a high CRR of 95.0\%, and non-adversarially trained methods show large robustness drops. ESD rises sharply from 26.5\% to 66.7\% (+40.2\%), and UCE increases from 18.5\% to 36.9\% (+18.4\%), revealing that suppression is far less reliable once prompts are adversarially optimized. By contrast, Receler maintains a low CRR of 14.8\%, demonstrating that adversarial training effectively mitigates such attacks.

\begin{figure*}[t]
    \centering
    \begin{subfigure}[t]{0.3\textwidth}
        \centering
        \includegraphics[width=\linewidth]{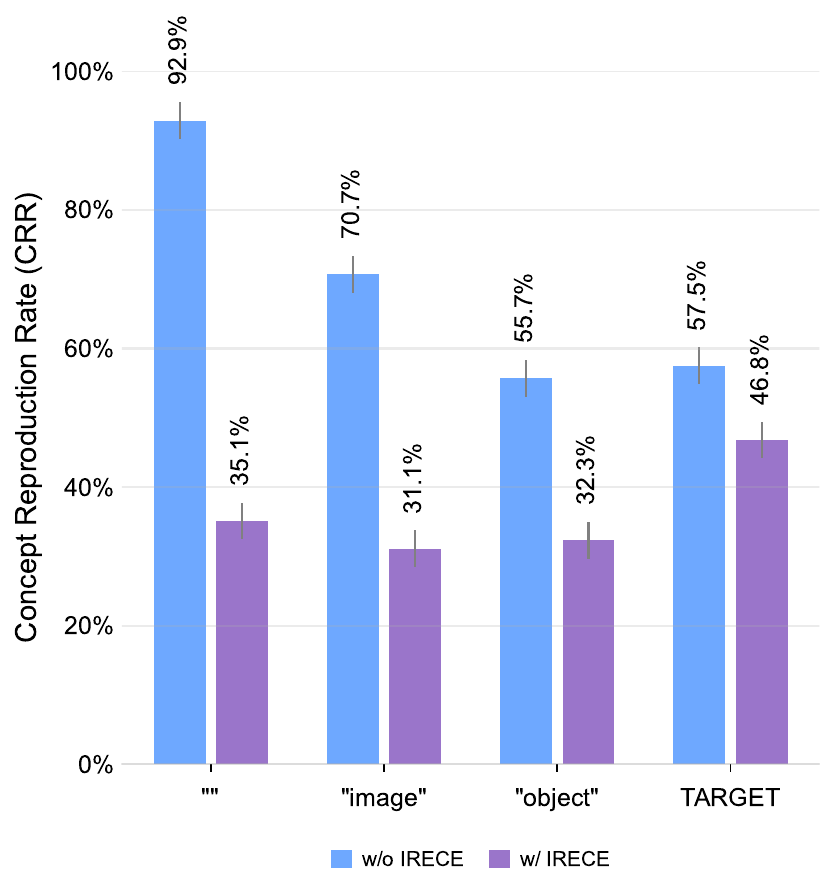}
        \caption{ESD with White-box Setting}
        \label{fig:irece_esd_white_box}
    \end{subfigure}
    \hfill
    \begin{subfigure}[t]{0.3\textwidth}
        \centering
        \includegraphics[width=\linewidth]{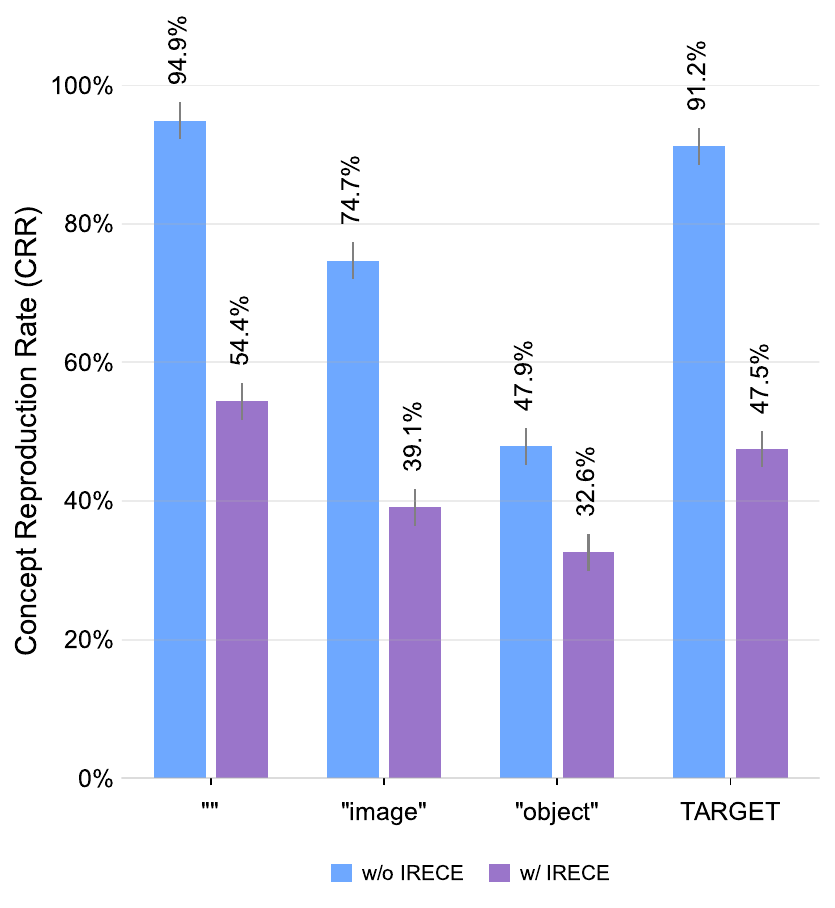}
        \caption{UCE with White-box Setting}
        \label{fig:irece_uce_white_box}
    \end{subfigure}
    \hfill
    \begin{subfigure}[t]{0.3\textwidth}
        \centering
        \includegraphics[width=\linewidth]{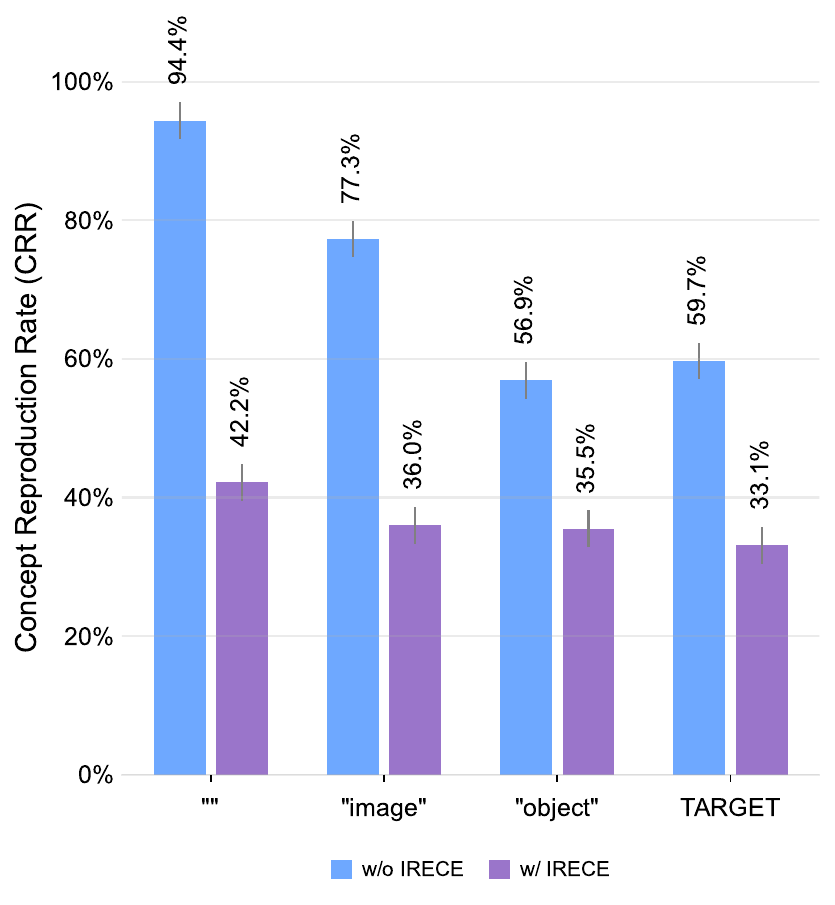}
        \caption{Receler with White-box Setting}
        \label{fig:irece_receler_white_box}
    \end{subfigure}
    \centering
    \begin{subfigure}[t]{0.3\textwidth}
        \centering
        \includegraphics[width=\linewidth]{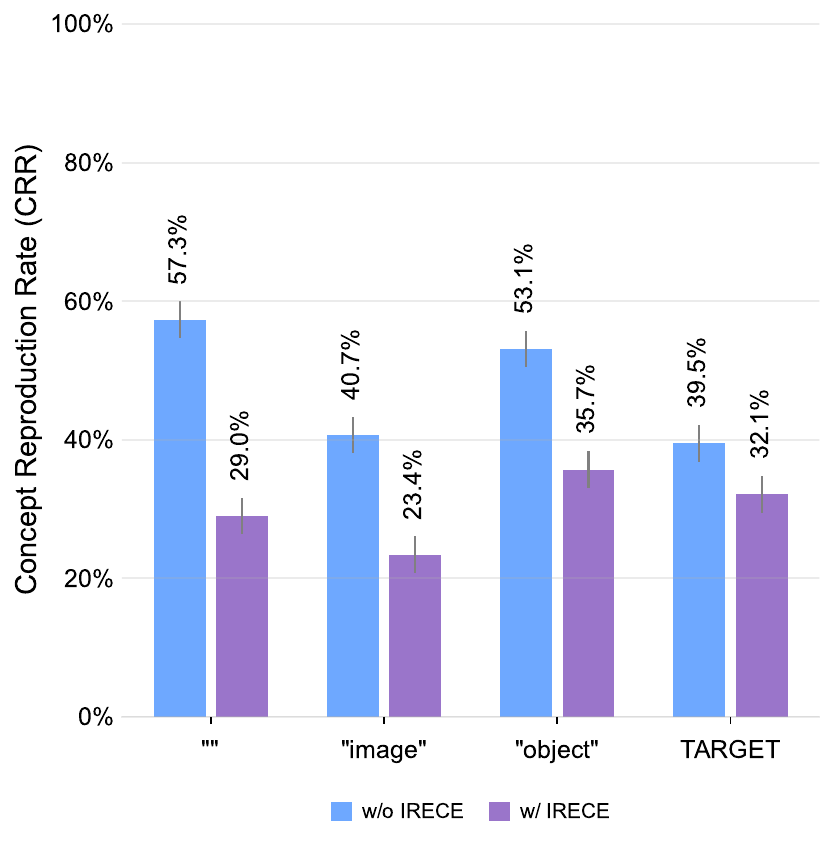}
        \caption{ESD with Black-box Setting}
        \label{fig:irece_esd_black_box}
    \end{subfigure}
    \hfill
    \begin{subfigure}[t]{0.3\textwidth}
        \centering
        \includegraphics[width=\linewidth]{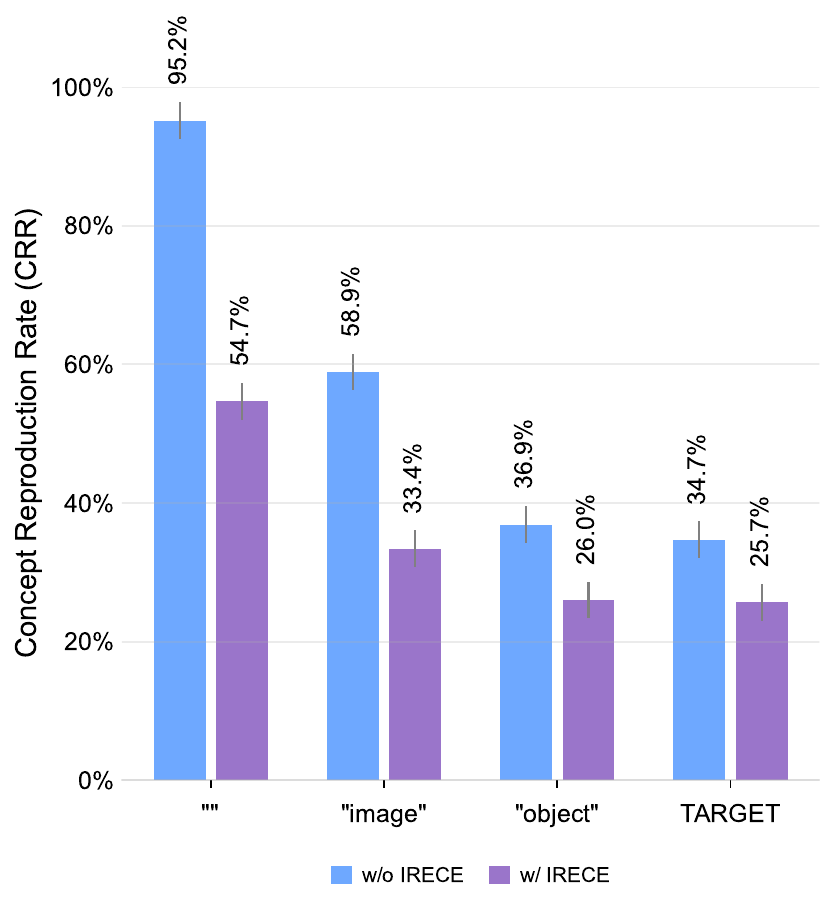}
        \caption{UCE with Black-box Setting}
        \label{fig:irece_uce_black_box}
    \end{subfigure}
    \hfill
    \begin{subfigure}[t]{0.3\textwidth}
        \centering
        \includegraphics[width=\linewidth]{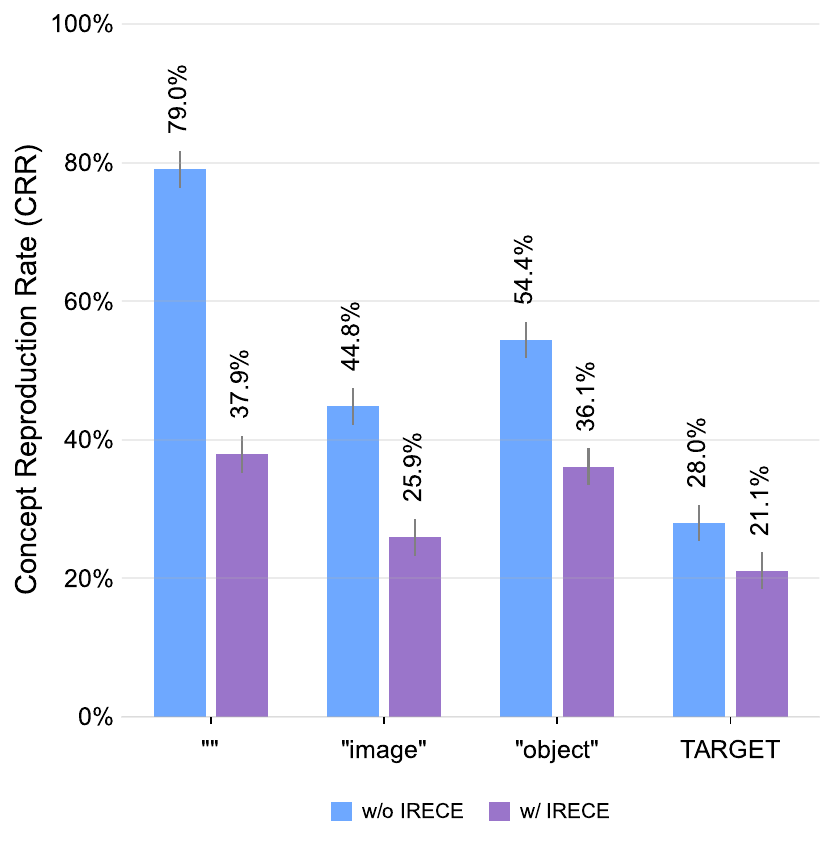}
        \caption{Receler with Black-box Setting}
        \label{fig:irece_receler_black_box}
    \end{subfigure}
    \caption{
    \textbf{Comparison of Concept Reproduction Rate (CRR) with and without IRECE under latent-inversion evaluation.} Subfigures present results for three representative concept-erasure methods under white-box and black-box settings. The horizontal axis groups correspond to different prompt types (``", ``image", ``object" and \texttt{TARGET}), and bar colors indicate whether IRECE is applied.
    }
    \label{fig:irece}
    \vspace{-10pt}
\end{figure*}

\subsubsection{Learned Embedding Evaluation}
\begin{figure}[h]
    \centering
    \includegraphics[width=\linewidth]{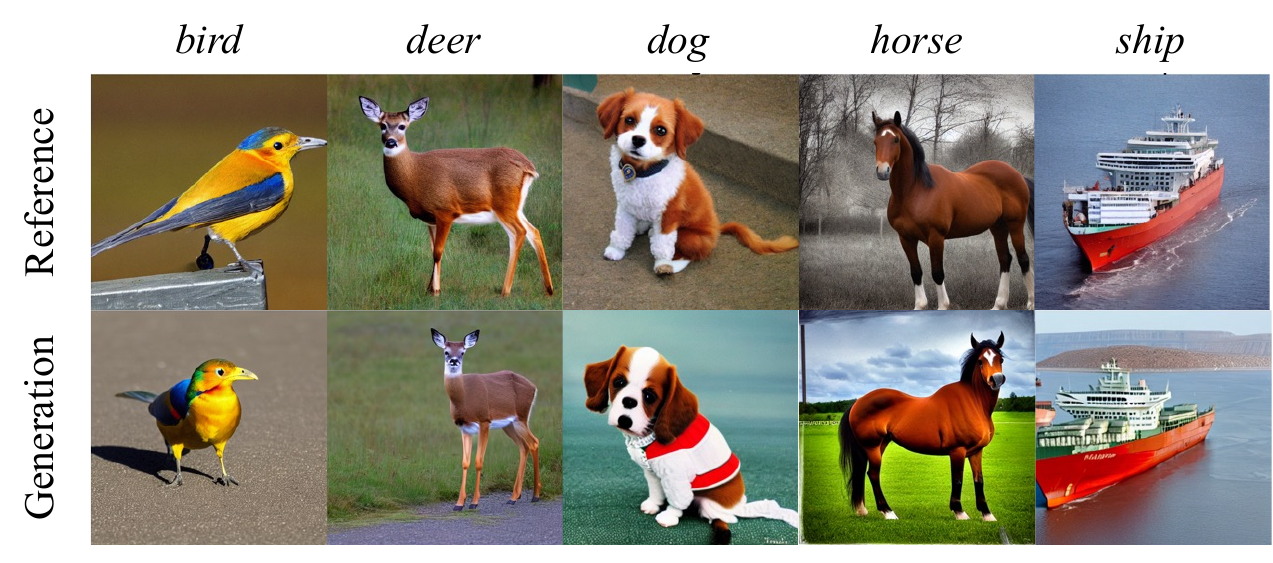}
    \caption{
    Qualitative results of generated images from concept-erased diffusion models under the black-box setting with perturbed reference images in the learned embedding evaluation.
    \vspace{-10pt}
    }
    \label{fig:learn_emb_black_box_perturb}
\end{figure}

We evaluate concept erasure methods on the SD-TI dataset, with results shown in Figure~\ref{fig:text_prompt_and_learned_emb_eval}(b). In the white-box setting, where embeddings are learned directly on the erased model, all methods exhibit a sharp rise in CRR compared to their performance in the text prompt evaluation. ESD increases from 26.5\% to 91.1\% (+64.6\%), UCE from 18.5\% to 90.4\% (+71.9\%), and even Receler, despite adversarial training, rises from 15.0\% to 56.0\% (+41.0\%). These results indicate that learned embeddings substantially weaken the suppression capability of erased models. Since white-box access is rarely realistic, we further evaluate in the black-box setting, where embeddings are learned on the standard model but tested on the erased one. Here, CRR drops considerably: ESD decreases to 41.2\%, UCE to 35.7\%, and Receler to 12.4\%. Notably, Receler even falls below its text prompt baseline, while ESD and UCE remain higher than their text prompt evaluations, confirming that embeddings continue to pose a challenge. To further strengthen the evaluation procedure, we introduce small perturbations (with a per-pixel perturbation budget of 8) to the reference images when training embeddings in the black-box setting. This strategy encourages semantic drift in the learned embeddings, making suppression even harder. CRR rises sharply again: ESD increases from 41.2\% to 74.0\% (+32.8\%), UCE from 35.7\% to 54.2\% (+18.5\%), while Receler shows only a marginal increase, reinforcing its relative robustness. Our analysis reveals that learned embeddings substantially reduces the effectiveness of erasure, and perturbations further amplify this effect, with only Receler retaining partial robustness. Figure~\ref{fig:learn_emb_black_box_perturb} shows that, in the black-box setting with perturbation, the target concept is successfully regenerated.

\begin{figure*}[t]
    \centering
    \includegraphics[width=\linewidth]{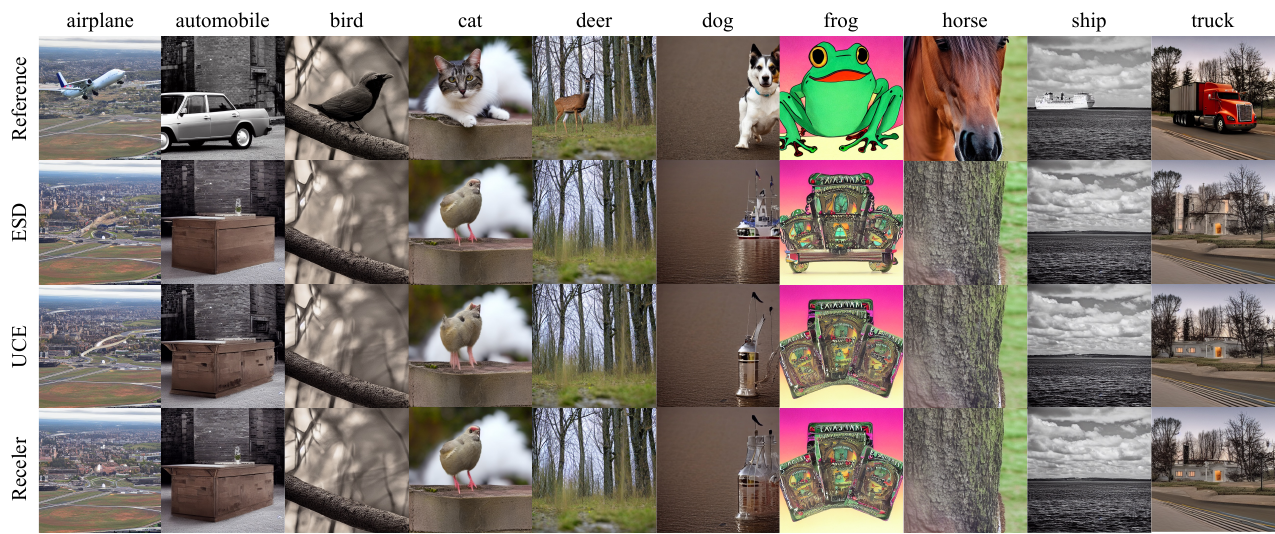}
    \caption{
    \textbf{Comparison of erased models with plug-and-play IRECE module across 10 target concepts.} These results are generated under the white-box setting using the unconditional prompt. The first column shows the reference images used to obtain the inverted latents. The remaining columns present the outputs from different concept erasure methods. After applying IRECE, the corresponding target concept is effectively removed from all outputs, while the rest of each image remains visually similar to its reference.
    \vspace{-10pt}
    }
    \label{fig:irece_vis}
\end{figure*}

\begin{figure}[h]
    \centering
    \includegraphics[width=\linewidth]{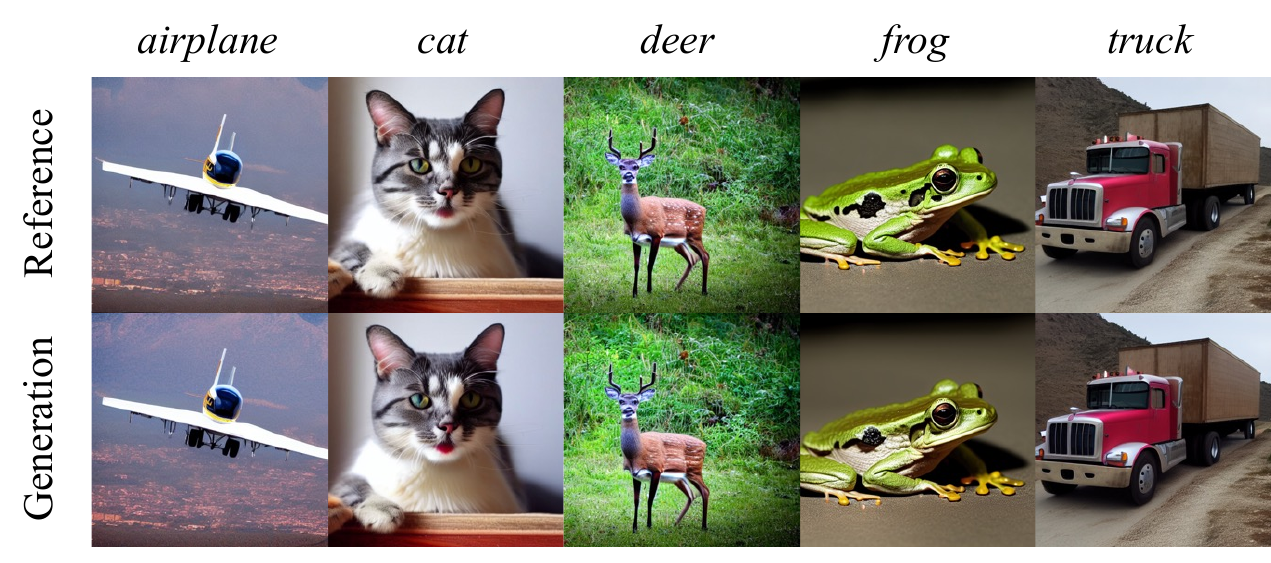}
    \caption{
    Qualitative results for generated images from concept-erased diffusion models with unconditional prompt under the black-box latent inversion evaluation.
    \vspace{-15pt}
    }
    \label{fig:latent_inv_black_box_null}
\end{figure}

\subsubsection{Latent Inversion Evaluation}

We evaluate concept erasure methods on the SD-LatentInv dataset, with results presented in Figure~\ref{fig:latent_eval}. In the white-box setting, latent inversion is highly effective in bypassing erasure. The null text prompt drives CRR above 92\% for all methods, making it the most challenging setting. The ``image" strategy also yields consistently strong results, with CRR exceeding 70.0\% across all methods. Even the \texttt{TARGET} strategy surpasses 50.0\%, showing that inversion substantially weakens suppression regardless of prompt type. In the black-box setting, CRR remains higher than the text prompt baselines across all strategies. Even the weakest case, \texttt{TARGET}, shows an increase of around 13.0\% compared to text prompt evaluation. The unconditional prompt remains the strongest: ESD rises from 26.5\% to 57.3\% (+30.8\%), UCE from 18.5\% to 95.2\% (+76.7\%), and Receler from 15.0\% to 79.0\% (+64.0\%). Our analysis reveals that latent inversion exposes the most severe vulnerability. By directly initializing the generative process from concept-containing latents, erased concepts frequently re-emerge, and even adversarially trained methods like Receler fail to maintain robustness. This underscores the fundamental limitation of current text-driven defenses.

Figure~\ref{fig:latent_inv_black_box_null} shows that, in the black-box setting with the null text prompt, the target concept is successfully regenerated, and the generated images maintain reconstruction quality comparable to the reference images.

\subsection{Inference-time Robustness Enhancement for Concept Erasure (IRECE)}

Quantitative results in Figure~\ref{fig:irece} show that IRECE consistently improves robustness across all settings. The largest gains occur under the most vulnerable configuration, white-box latent inversion with the null text prompt, where CRR decreases by more than 40\% for all methods. Specifically,
CRR is reduced from 92.9\% to 35.1\% for ESD (-57.8\%), from 94.4\% to 42.2\% for Receler (-52.2\%), and from 94.9\% to 54.4\% for UCE (-40.5\%). These results indicate that IRECE effectively mitigates the weakest cases of concept erasure, restoring robustness even under the most challenging settings. We include per-class CRR results in Tables 6 and 7 of the Appendix, along with the ablation study in Section C.5. Qualitative results in Figure~\ref{fig:irece_vis} provide further evidence. For several classes (\textit{airplane}, \textit{bird}, \textit{deer}, \textit{ship}, and \textit{truck}), the target concept is almost entirely removed, while for others (\textit{automobile}, \textit{cat}, \textit{dog}, \textit{frog}, and \textit{horse}), it is replaced with alternative content. In both cases, non-target regions remain largely intact, and transitions between modified and unmodified areas appear visually coherent. These examples confirm that IRECE not only strengthens robustness but also enables controlled object removal or replacement in practice.

\section{Conclusion}
In this paper, we introduce \bench{} to assess the robustness of concept erasure methods in diffusion models. While existing approaches perform well on text prompt evaluation, they degrade substantially when tested with learned embeddings and nearly fail under image latent inversion, exposing critical vulnerabilities. To address these gaps, we proposed IRECE, a plug-and-play inference-time module that integrates seamlessly with erased models without retraining. Experiments demonstrate that IRECE consistently restores robustness, reducing Concept Reproduction Rate by up to 40\% under the most challenging white-box latent inversion setting, while maintaining visual quality. This establishes IRECE as an effective defense against multimodal attacks on erased models. Beyond robustness, IRECE also enables broader applications, such as targeted object removal and replacement, underscoring the importance of reliable concept erasure in generative models.

\section*{Acknowledgements}

This research is supported by the National Science and Technology Council (NSTC) under the grant of NSTC-114-2634-F-002-004, NSTC-114-2634-F-001-001-MBK, NSTC-113-2634-F-002-008, NSTC-114-2221-E-001-016 and NSTC-114-2221-E-001-004, NSTC-114-2221-E-002-182-MY3, NSTC-113-2221-E-002-201 and Academia Sinica under the grant of AS-CDA-110-M09 and AS-IAIA-114-M10.

\clearpage
{
    \small
    \bibliographystyle{ieeenat_fullname}
    \bibliography{ref}
}

\clearpage
\renewcommand\thesection{\Alph{section}}
\setcounter{section}{0}

\twocolumn[
    \vspace{1cm}
    \begin{center}
        \Large{\textbf{M-ErasureBench: A Comprehensive Multimodal Evaluation Benchmark for
        Concept Erasure in Diffusion Models Supplementary Material}}
    \end{center}
    \vspace{1cm}
]

\section{IRECE Algorithm}
We present the detailed algorithm of IRECE in Algorithm~\ref{algo:irece}.

\begin{algorithm}[H]
    \caption{IRECE} \label{algo:irece}
    \begin{algorithmic}[1] 
        \STATE \textbf{Input:} 
        Sample prompt $\bm{c}_{\text{sam}}$, 
        Target prompt $\bm{c}_{\text{tgt}}$, 
        Initial latent $\bm{x}_T$, 
        Noise schedule $\{\bar{\alpha}_t\}_{t=1}^T$, 
        Intervention step $t^*$, 
        Concept localization threshold $\tau$, 
        Standard model $\theta_{\text{std}}$, 
        Erased model $\theta_{\text{era}}$.
        \FOR{$t = T$ to $1$}
            \IF{$t = t^*$}
                \STATE Extract cross-attention maps from each layer $\ell$ of model $\theta$ (white-box: $\theta_{\text{era}}$, black-box: $\theta_{\text{std}}$):
                \[
                    \bm{A}^\ell \gets A_\text{cross}^\ell(\bm{x}_t, \bm{c}_{\text{tgt}}; \theta)
                \]
                \STATE Aggregate maps after upsampling:
                \[
                    \bm{A} \gets \sum_{\ell=1}^L \operatorname{Upsample}(\bm{A}^\ell)
                \]
                \STATE Construct binary mask:
                \[
                \bm{M}(i,j) \gets 
                \begin{cases}
                1, & \bm{A}(i,j) \geq \tau \\
                0, & \text{otherwise}
                \end{cases}
                \]
                \STATE Perturb target regions with Gaussian noise $\bm{\xi}_t \sim \mathcal{N}(\bm{0},\bm{I})$:
                \[
                    \bm{x}_t \gets (\bm{1}-\bm{M}) \odot \bm{x}_t + \bm{M} \odot \bm{\xi}_t
                \]
            \ENDIF
            \STATE Perform the denoising step with the erased model:
            \[
                \bm{x}_{t-1} \gets \operatorname{DDIMStep}(\bm{x}_t, \bm{c}_\text{sam}, t, \theta_\text{era})
            \]
        \ENDFOR
        \STATE \textbf{Output:} Generated output $\bm{x}_0$.
    \end{algorithmic}
\end{algorithm}

\section{Implementation Details}

\paragraph{Sampling.}
All sampling procedures are conducted using the \texttt{DDIMScheduler} in diffuser~\cite{platen2022diffusers}, with the guidance scale set to 7.5 and the number of inference fixed at 50.

\paragraph{Prompt Configurations.}
Table~\ref{tab:prompt_pair} presents the descriptions of the four prompt configurations used in the latent inversion evaluation.

\begin{table}[h]
    \centering
    \resizebox{\linewidth}{!}{%
    \begin{tabular}{cl}
    \toprule
    Prompt & Description \\
    \midrule
    \multirow{2}{*}{``"} & Null text (unconditional generation with no \\
    ~ & textual guidance). \\
    \multirow{2}{*}{``image"} & Generic placeholder describing the input \\
    ~ & image without specifying any object. \\
    \multirow{2}{*}{``object"} & Coarse reference to the foreground object \\
    ~ & without explicitly naming it. \\
    \multirow{2}{*}{\texttt{TARGET}} & Explicitly naming the target concept intended \\
    ~ & for erasure. \\
    \bottomrule
    \end{tabular}
    }
    \caption{Prompt configurations for latent inversion evaluation.}
    \vspace{-10pt}
    \label{tab:prompt_pair}
\end{table}

\paragraph{Parameters of IRECE.}
For robustness enhancement, the concept localization threshold $\tau$ is set to 0.4 and the intervention timestep $t^*$ at 781.

\section{More Experimental Results}

\subsection{Text Prompt Evaluation}

We report per-class results for both text prompt and adversarial prompt evaluations in Table~\ref{tab:text_prompt_eval_per_class}. Across the ten categories, \textit{automobile} consistently exhibits the highest CRR under all methods, indicating that erasure is less effective for this class. A plausible reason is the large intra-class diversity of automobiles, which makes the concept harder to suppress. Despite this challenge, all methods still reduce CRR by at least 39\% for \textit{automobile}, confirming that suppression remains non-trivial but effective to some extent.

\begin{table*}[h]
\centering
\begin{tabular}{l*{11}{c}}
\toprule
Methods & \textit{airplane} & \textit{automobile} & \textit{bird} & \textit{cat} & \textit{deer} & \textit{dog} & \textit{frog} & \textit{horse} & \textit{ship} & \textit{truck} & \textbf{Avg.} \\
\midrule
\multicolumn{12}{c}{\textit{Text Prompt}} \\
\midrule
SD v1.4~\cite{rombach2022ldm} & \highlight 94.67 & \highlight 98.67 & \highlight 92.67 & \highlight 96.00 & \highlight 99.33 & \highlight 98.67 & \highlight 92.00 & \highlight 98.67 & \highlight 94.67 & \highlight 96.00 & \highlightavg{\textbf{96.14}} \\
ESD~\cite{gandikota2023esd} & 27.33 & \highlight 59.33 & 10.00 & 31.33 & 16.67 & 22.67 & 19.33 & 18.67 & 32.00 & 27.33 & \textbf{26.47} \\
UCE~\cite{gandikota2024uce} & 36.00 & 44.00 & 3.33  & 9.33  & 7.33  & 6.00  & 14.67 & 5.33  & 30.67 & 28.00 & \textbf{18.47} \\
Receler~\cite{huang2024receler} & 6.67  & \highlight 50.67 & 2.67  & 6.67  & 5.33  & 2.00  & 25.33 & 8.67 & 26.67 & 15.33 & \textbf{15.00} \\
\midrule
\multicolumn{12}{c}{\textit{Adversarial Prompt}} \\
\midrule
SD v1.4~\cite{rombach2022ldm} & \highlight 85.33 & \highlight 95.56 & \highlight 94.67 & \highlight 95.56 & \highlight 100.00 & \highlight 97.33 & \highlight 99.33 & \highlight 96.00 & \highlight 92.67 & \highlight 93.33 & \highlightavg{\textbf{94.98}} \\
ESD~\cite{gandikota2023esd} & \highlight 74.67 & \highlight 95.56 & \highlight 69.33 & \highlight 52.14 & \highlight 50.00 & \highlight 79.33 & 32.00 & 47.33 & \highlight 82.67 & \highlight 84.17 & \highlightavg{\textbf{66.72}} \\
UCE~\cite{gandikota2024uce} & \highlight 70.67 & \highlight 72.67 & 30.00  & 14.67  & 9.33  & 22.67  & 35.33 & 18.67  & 43.33 & \highlight 52.00 & \textbf{36.93} \\
Receler~\cite{huang2024receler} & 8.67  & \highlight 78.89 & 4.00  & 0.00  & 0.00  & 2.00  & 11.33 & 0.00  & 34.67 & 8.33 & \textbf{14.79} \\
\bottomrule
\end{tabular}
\caption{
Concept Reproduction Rate (CRR) of concept erasure methods on \textbf{text} and \textbf{adversarial prompts}, reported per class. \highlighttext{Orange} marks classes with CRR $>$ 50\%, and \highlightavgtext{red} marks methods with average CRR $>$ 50\%.
}
\label{tab:text_prompt_eval_per_class}
\end{table*}

\begin{table*}[h]
\centering
\resizebox{\textwidth}{!}{
\begin{tabular}{llrrrrrrrrrrr}
\toprule
Methods & Settings & \textit{airplane} & \textit{automobile} & \textit{bird} & \textit{cat} & \textit{deer} & \textit{dog} & \textit{frog} & \textit{horse} & \textit{ship} & \textit{truck} & \textbf{Avg.} \\
\midrule
\multirow{4}{*}{ESD~\cite{gandikota2023esd}}
 & Text prompt & 27.33 & \highlight 59.33 & 10.00 & 31.33 & 16.67 & 22.67 & 19.33 & 18.67 & 32.00 & 27.33 & \textbf{26.47} \\
 & White-box & \highlight 78.00 & \highlight 97.33 & \highlight 93.33 & \highlight 98.00 & \highlight 94.67 & \highlight 96.67 & \highlight 82.67 & \highlight 84.67 & \highlight 96.67 & \highlight 88.67 & \highlightavg{\textbf{91.07}} \\
 & Black-box & 44.67 & \highlight 82.67 & 15.33 & 37.33 & 30.00 & 31.33 & 25.33 & 18.00 & \highlight 86.67 & 40.67 & \textbf{41.20} \\
 & Black-box w/ perturb. & \highlight 74.67 & \highlight 97.33 & \highlight 88.00 & \highlight 81.33 & 39.33 & \highlight 80.67 & \highlight 52.67 & \highlight 66.00 & \highlight 93.33 & \highlight 66.67 & \highlightavg{\textbf{74.00}} \\
\midrule
\multirow{4}{*}{UCE~\cite{gandikota2024uce}} 
 & Text prompt & 36.00 & 44.00 & 3.33 & 9.33 & 7.33 & 6.00 & 14.67 & 5.33 & 30.67 & 28.00 & \textbf{18.47} \\
 & White-box & \highlight 70.67 & \highlight 98.67 & \highlight 92.67 & \highlight 96.00 & \highlight  93.33 & \highlight 84.00 & \highlight 90.67 & \highlight 99.33 & \highlight 95.33 & \highlight 83.33 & \highlightavg{\textbf{90.40}} \\
 & Black-box& 7.33 & \highlight 92.00 & 8.00 & \highlight 62.00 & 23.33 & 22.67 & 16.67 & 8.00 & \highlight 70.00 & 46.67 & \textbf{35.67} \\
 & Black-box w/ perturb. & 34.00 & \highlight 86.67 & \highlight 54.67 & \highlight 93.33 & \highlight 65.33 & 48.00 & 29.33 & 14.00 & \highlight 59.33 & \highlight 57.33 & \highlightavg{\textbf{54.20}} \\
\midrule
\multirow{4}{*}{Receler~\cite{huang2024receler}} 
 & Text prompt & 6.67 & \highlight 50.67 & 2.67 & 6.67 & 5.33 & 2.00 & 25.33 & 8.67 & 26.67 & 15.33 & \textbf{15.00} \\
 & White-box & \highlight 50.00 & \highlight 89.33 & 18.00 & 39.33 & 41.33 & \highlight 92.00 & 34.67 & 22.67 & \highlight 82.67 & \highlight 90.00 & \highlightavg{\textbf{56.00}} \\
 & Black-box& 8.67 & \highlight 64.00 & 0.67 & 0.00 & 1.33 & 1.33 & 16.67 & 0.00 & 10.67 & 20.67 & \textbf{12.40} \\
 & Black-box w/ perturb. & 12.00 & \highlight 78.00 & 2.00 & 2.67 & 1.33 & 3.33 & 12.67 & 2.67 & 24.67 & 11.33 & \textbf{15.07} \\
\bottomrule
\end{tabular}
}
\caption{
Concept Reproduction Rate (CRR) of concept-erasure methods in \textbf{learned embedding evaluation}, reported per class. \highlighttext{Orange} marks classes with CRR $>$ 50\%, and \highlightavgtext{red} marks methods with average CRR $>$ 50\%.
}
\label{tab:learned_embedding_eval_per_class}
\vspace{-10pt}
\end{table*}

\subsection{Learned Embedding Evaluation}
We present per-class results for learned embedding evaluation in Table~\ref{tab:learned_embedding_eval_per_class}. Compared to the text prompt baseline, CRR under the white-box setting rises above 50\% for many categories across all methods, indicating that learned embeddings substantially reduces the effectiveness of erasure. In the black-box setting, CRR remains below 50\% for most categories, but introducing perturbations substantially increases CRR. For example, ESD and UCE exceed 50\% CRR in classes such as \textit{bird}, \textit{cat}, \textit{deer}, and \textit{truck}, highlighting that even black-box settings can become highly effective when enhanced with perturbations.

\subsection{Latent Inversion Evaluation}
We report per-class results for latent inversion evaluation in Table~\ref{tab:latent_inversion_eval_per_class}. 
In the white-box setting, concept erasure methods show consistently high CRR: with prompts such as ``'' and ``image'', nearly all categories exceed 50\%, while only \texttt{TARGET} achieves CRR below 50\% in a few cases (e.g., \textit{deer}, \textit{horse}). In the black-box setting, the unconditional prompt ``'' still drives CRR above 50\% for most categories, underscoring the vulnerability of erased models under latent inversion evaluation.

\begin{figure}[h]
    \centering
    \includegraphics[width=\linewidth]{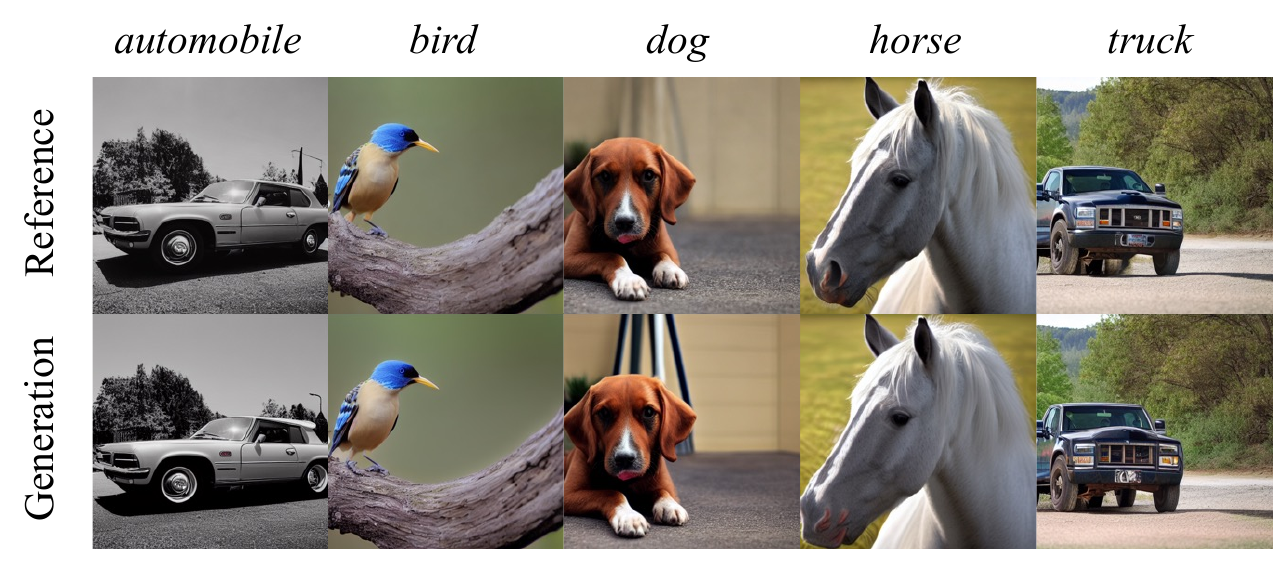}
    \caption{
    Qualitative results for generated images with ``image" prompt under the black-box latent inversion evaluation.
    }
    \label{fig:latent_inv_white_box_image}
    \vspace{-5pt}
\end{figure}

\begin{figure}[h]
    \centering
    \includegraphics[width=\linewidth]{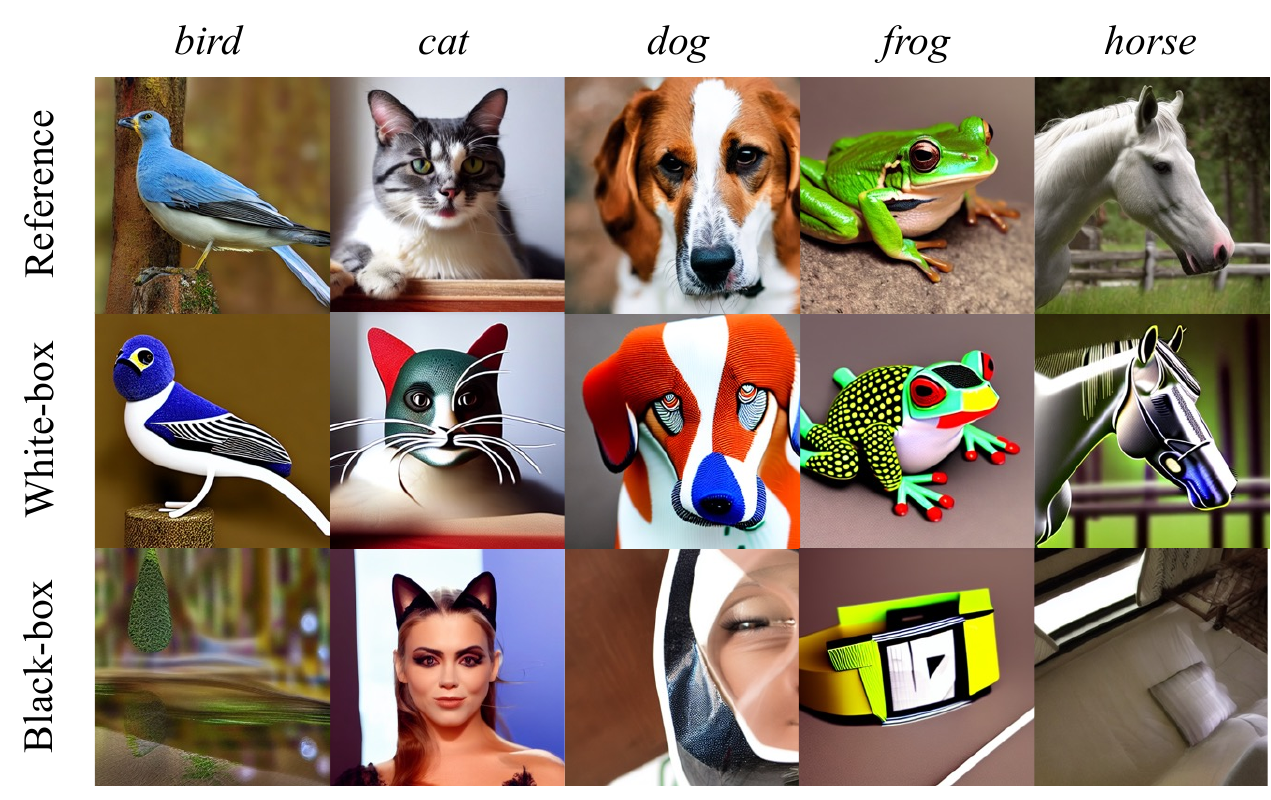}
    \caption{
    Qualitative results for \texttt{TARGET} prompt generations under white-box and black-box latent inversion evaluation.}
    \label{fig:latent-target}
\end{figure}

Figure~\ref{fig:latent_inv_white_box_image} shows that the ``image'' prompt strategy also performs strongly under white-box access, successfully capturing the overall semantics in most cases. In contrast, the \texttt{TARGET} strategy exhibits markedly different behaviors across access settings. As shown in Figure~\ref{fig:latent-target}, under black-box access it often achieves effective concept removal, producing outputs that deviate substantially from the original semantics. In the white-box case, however, the generated images continue to depict the target concept, albeit with noticeable disruptions in the corresponding regions.

\subsection{(Surrogate-based) Black-box with Different Backbones}

We further investigate how backbone discrepancies between the surrogate and erased models affect the effectiveness of concept erasure. Since the null text prompt in the white-box setting yields the most prominent performance (Table~\ref{tab:latent_inversion_eval_per_class}), we mainly adopt it as our case study for \textbf{latent inversion evaluation}. Specifically, we generate inverted latents via DDIM inversion using SD v1.5, while all concept-erasure methods are evaluated on an erased model based on SD v1.4. As shown in Table~\ref{tab:learned_embedding_different_backbone}, and in reference to the null-text prompt results in Table~\ref{tab:latent_inversion_eval_per_class}, when the surrogate and erased models use different backbones, the CRRs are lower than those in the white-box setting; meanwhile, they remains consistently higher than those in the corresponding black-box setting where both models share the same backbone. This is because the backbone mismatch causes the surrogate’s representation space to deviate from that of the erased model, producing latents falling outside the regions where the erased model was trained to suppress the target concept.

\subsection{Inference-time Robustness Enhancement for Concept Erasure (IRECE)}

We report the detailed per-class CRR results before and after applying IRECE for three concept erasure methods: ESD~\cite{gandikota2023esd}, UCE~\cite{gandikota2024uce}, and Receler~\cite{huang2024receler}. White-box results are shown in Table~\ref{tab:irece_white}, and those under the black-box setting are provided in Table~\ref{tab:irece_black}. Across most classes and prompt strategies, IRECE achieves a substantial reduction in CRR, indicating improved robustness against concept re-emergence.

\begin{figure*}[t]
    \centering
    \includegraphics[width=\linewidth]{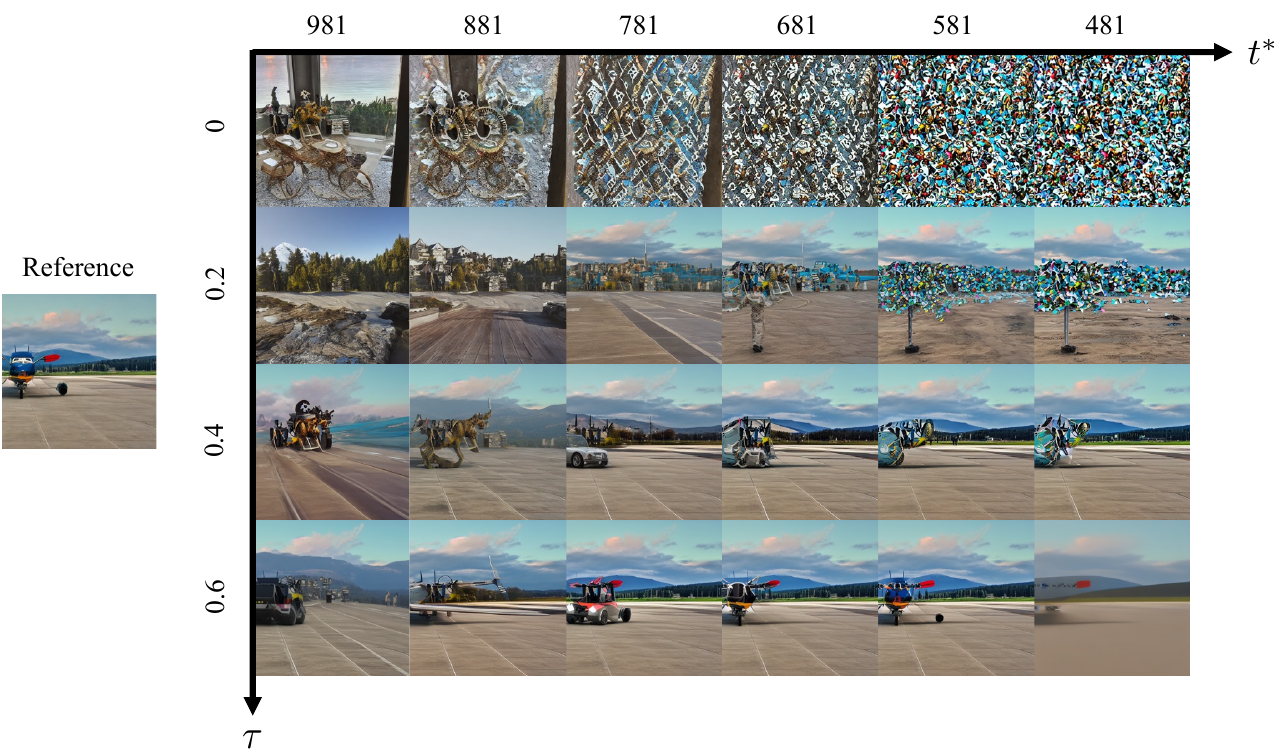}
    \caption{
    \textbf{Effect of intervention timestep and concept localization threshold on IRECE.}
    Columns correspond to intervention timesteps $t^*$ (decreasing left to right), and rows to concept localization thresholds $\tau$ (increasing top to bottom).
    }
    \label{fig:irece_ablation}
\end{figure*}

IRECE introduces two tunable hyperparameters: the intervention timestep $t^*$ and the concept localization threshold $\tau$. Their effects are shown in Figure~\ref{fig:irece_ablation}.

\paragraph{Intervention Timestep.}
This parameter specifies when the erasure is applied during the diffusion process. Applying it too early disrupts the overall image structure, as the latent representation is not yet sufficiently developed. Applying it too late leaves too few denoising steps, often leading to blending artifacts and incomplete suppression.

\paragraph{Concept Localization Threshold.}
This parameter controls the aggressiveness of erasure. A lower $\tau$ produces broader masks that may unintentionally remove nearby non-target regions, while a higher $\tau$ results in tighter masks that risk leaving traces of the target concept. Its role is analogous to the scale coefficient in Receler~\cite{huang2024receler}, which adjusts the trade-off between erasure strength and image fidelity.

\begin{table*}[h]
\centering
\resizebox{\textwidth}{!}{%
\begin{tabular}{llrrrrrrrrrrrr}
\toprule
Methods & Prompt Strategy & \textit{airplane} & \textit{automobile} & \textit{bird} & \textit{cat} & \textit{deer} & \textit{dog} & \textit{frog} & \textit{horse} & \textit{ship} & \textit{truck} & \textbf{Avg.} \\
\midrule
\multicolumn{13}{c}{\textit{White-box Setting}} \\
\midrule
\multirow{5}{*}{ESD~\cite{gandikota2023esd}} 
& Text prompt & 27.33 & \highlight 59.33 & 10.00 & 31.33 & 16.67 & 22.67 & 19.33 & 18.67 & 32.00 & 27.33 & \textbf{26.47} \\
& ``" & \highlight 88.67 & \highlight 98.00 & \highlight 89.33 & \highlight 94.67 & \highlight 94.67 & \highlight 98.00 & \highlight 92.67 & \highlight 95.33 & \highlight 86.67 & \highlight 90.67 & \highlightavg{\textbf{92.87}} \\
& ``image" & \highlight 71.33 & \highlight 86.00 & \highlight 76.00 & \highlight 68.67 & \highlight 57.33 & \highlight 73.33 & \highlight 78.67 & \highlight 65.33 & \highlight 70.67 & \highlight 59.33 & \highlightavg{\textbf{70.67}} \\
& ``object" & \highlight 72.67 & \highlight 80.67 & \highlight 52.00 & 48.67 & 38.67 & 39.33 & \highlight 68.00 & 48.00 & \highlight 60.00 & 48.67 & \highlightavg{\textbf{55.67}} \\
& \texttt{TARGET} & \highlight 68.00 & \highlight 86.67 & \highlight 52.00 & \highlight 50.67 & 35.33 & \highlight 58.67 & \highlight 58.67 & \highlight 55.33 & \highlight 63.33 & 46.00 & \highlightavg{\textbf{57.47}} \\
\midrule
\multirow{5}{*}{UCE~\cite{gandikota2024uce}} 
& Text prompt & 36.00 & 44.00 & 3.33 & 9.33 & 7.33 & 6.00 & 14.67 & 5.33 & 30.67 & 28.00 & \textbf{18.47} \\
& ``" & \highlight 92.00 & \highlight 98.00 & \highlight 91.33 & \highlight 96.67 & \highlight 98.00 & \highlight 98.67 & \highlight 92.67 & \highlight 97.33 & \highlight 89.33 & \highlight 95.33 & \highlightavg{\textbf{94.93}} \\
& ``image" & \highlight 70.00 & \highlight 77.33 & \highlight 83.33 & \highlight 73.33 & \highlight 75.33 & \highlight 62.00 & \highlight 76.67 & \highlight 78.67 & \highlight 73.33 & \highlight 76.67 & \highlightavg{\textbf{74.67}} \\
& ``object" & \highlight 60.00 & \highlight 75.33 & 46.67 & 35.33 & 30.67 & 29.33 & \highlight 74.00 & 37.33 & 49.33 & 41.33 & \textbf{47.93} \\
& \texttt{TARGET} & \highlight 84.67 & \highlight 96.67 & \highlight 90.67 & \highlight 94.67 & \highlight 86.67 & \highlight 94.67 & \highlight 91.33 & \highlight 92.67 & \highlight 88.00 & \highlight 92.00 & \highlightavg{\textbf{91.20}} \\
\midrule
\multirow{5}{*}{Receler~\cite{huang2024receler}} 
& Text prompt & 6.67 & \highlight 50.67 & 2.67 & 6.67 & 5.33 & 2.00 & 25.33 & 8.67 & 26.67 & 15.33 & \textbf{15.00} \\
& ``" & \highlight 91.33 & \highlight 100.00 & \highlight 90.67 & \highlight 94.67 & \highlight 97.33 & \highlight 98.67 & \highlight 92.00 & \highlight 96.67 & \highlight 88.00 & \highlight 94.67 & \highlightavg{\textbf{94.40}} \\
& ``image" & \highlight 72.67 & \highlight 84.67 & \highlight 82.00 & \highlight 74.67 & \highlight 68.67 & \highlight 82.00 & \highlight 80.67 & \highlight 78.00 & \highlight 68.67 & \highlight 80.67 & \highlightavg{\textbf{77.27}} \\
& ``object" & \highlight 67.33 & \highlight 72.67 & \highlight 54.00 & \highlight 50.00 & 48.00 & 38.67 & \highlight 77.33 & 42.67 & \highlight 63.33 & \highlight 54.67 & \highlightavg{\textbf{56.87}} \\
& \texttt{TARGET} & \highlight 61.33 & \highlight 91.33 & \highlight 60.67 & \highlight 70.00 & 26.67 & \highlight 62.00 & \highlight 80.67 & 34.67 & \highlight 54.00 & \highlight 56.00 & \highlightavg{\textbf{59.73}} \\
\midrule
\multicolumn{13}{c}{\textit{Black-box Setting}} \\
\midrule
\multirow{5}{*}{ESD~\cite{gandikota2023esd}} 
& Text prompt & 27.33 & \highlight 59.33 & 10.00 & 31.33 & 16.67 & 22.67 & 19.33 & 18.67 & 32.00 & 27.33 & \textbf{26.47} \\
& ``" & \highlight 71.33 & \highlight 81.33 & \highlight 56.00 & 48.67 & 34.00 & \highlight 58.00 & \highlight 81.33 & \highlight 58.00 & \highlight 54.67 & 30.00 & \highlightavg{\textbf{57.33}} \\
& ``image"  & \highlight 51.33 & \highlight 73.33 & 24.67 & 34.67 & 16.00 & 40.00 & \highlight 62.00 & 29.33 & 39.33 & 36.00 & \textbf{40.67} \\
& ``object" & \highlight 60.67 & \highlight 83.33 & 37.33 & 33.33 & 32.00 & 46.00 & \highlight 80.00 & 43.33 & \highlight 58.67 & \highlight 56.67 & \highlightavg{\textbf{53.13}} \\
& \texttt{TARGET} & 46.67 & \highlight 74.00 & 16.67 & 32.00 & 35.33 & 34.00 & \highlight 57.33 & 27.33 & 32.67 & 38.67 & \textbf{39.47} \\
\midrule
\multirow{5}{*}{UCE~\cite{gandikota2024uce}} 
& Text prompt  & 36.00 & 44.00 & 3.33 & 9.33 & 7.33 & 6.00 & 14.67 & 5.33 & 30.67 & 28.00 & \textbf{18.47} \\
& ``" & \highlight 91.33 & \highlight 98.00 & \highlight 92.67 & \highlight 96.67 & \highlight 98.00 & \highlight 98.67 & \highlight 93.33 & \highlight 97.33 & \highlight 90.00 & \highlight 96.00 & \highlightavg{\textbf{95.20}} \\
& ``image"  & \highlight 54.00 & \highlight 70.00 & \highlight 62.67 & \highlight 60.67 & \highlight 54.67 & 41.33 & \highlight 75.33 & 43.33 & 34.67 & \highlight 92.00 & \highlightavg{\textbf{58.87}} \\
& ``object" & 31.33 & \highlight 74.00 & 20.67 & 8.00 & 15.33 & 21.33 & \highlight 63.33 & 30.00 & 37.33 & \highlight 68.00 & \textbf{36.93} \\
& \texttt{TARGET} & 40.67 & \highlight 74.00 & 11.33 & 28.00 & 30.00 & 30.00 & \highlight 50.00 & 25.33 & 24.67 & 33.33 & \textbf{34.73} \\
\midrule
\multirow{5}{*}{Receler~\cite{huang2024receler}} 
& Text prompt  & 6.67 & \highlight 50.67 & 2.67 & 6.67 & 5.33 & 2.00 & 25.33 & 8.67 & 26.67 & 15.33 & \textbf{15.00} \\
& ``" & \highlight 84.00 & \highlight 87.33 & \highlight 82.00 & \highlight 68.00 & \highlight 66.67 & \highlight 82.67 & \highlight 84.00 & \highlight 84.67 & \highlight 77.33 & \highlight 73.33 & \highlightavg{\textbf{79.00}} \\
& ``image"  & 43.33 & \highlight 73.33 & 36.67 & 40.00 & 26.00 & \highlight 50.00 & \highlight 56.67 & 43.33 & 27.33 & \highlight 51.33 & \textbf{44.80} \\
& ``object" & 70.00 & \highlight 66.67 & 33.33 & 33.33 & 46.00 & 48.00 & \highlight 71.33 & 40.67 & \highlight 56.67 & \highlight 78.00 & \highlightavg{\textbf{54.40}} \\
& \texttt{TARGET} & 28.67 & \highlight 69.33 & 12.00 & 17.33 & 23.33 & 12.00 & \highlight 50.67 & 23.33 & 17.33 & 26.00 & \textbf{28.00} \\
\bottomrule
\end{tabular}
}
\caption{
Concept Reproduction Rate (CRR) of concept-erasure methods in \textbf{latent inversion evaluation}, reported per class. \highlighttext{Orange} marks classes with CRR $>$ 50\%, and \highlightavgtext{red} marks methods with average CRR $>$ 50\%.
}
\label{tab:latent_inversion_eval_per_class}
\end{table*}

\begin{table*}[h]
\centering
\resizebox{\textwidth}{!}{
\begin{tabular}{lrrrrrrrrrrr}
\toprule
Methods & \textit{airplane} & \textit{automobile} & \textit{bird} & \textit{cat} & \textit{deer} & \textit{dog} & \textit{frog} & \textit{horse} & \textit{ship} & \textit{truck} & \textbf{Avg.} \\
\midrule
ESD~\cite{gandikota2023esd} & \highlight 82.67 & \highlight 94.00 & \highlight 93.33 & \highlight 91.33 & \highlight 79.33 & \highlight 90.67 & \highlight 89.33 & \highlight 94.67 & \highlight 90.67 & \highlight 93.33 & \highlightavg{\textbf{89.93}} \\
UCE~\cite{gandikota2024uce} & \highlight 86.00 & \highlight 58.67 & \highlight 77.33 & \highlight 70.00 & \highlight 58.00 & \highlight 83.33 & \highlight 82.67 & \highlight 88.00 & \highlight 94.67 & \highlight 82.00 & \highlightavg{\textbf{78.07}} \\
Receler~\cite{huang2024receler} & \highlight 86.67 & \highlight 94.00 & \highlight 93.33 & \highlight 87.33 & \highlight 78.67 & \highlight 92.67 & \highlight 90.00 & \highlight 95.33 & \highlight 90.67 & \highlight 96.00 & \highlightavg{\textbf{90.47}} \\
\bottomrule
\end{tabular}
}
\caption{Concept Reproduction Rate (CRR) of concept-erasure methods in \textbf{latent inversion evaluation with different backbone}, reported per class. \highlighttext{Orange} marks classes with CRR $>$ 50\%, and \highlightavgtext{red} highlights methods with average CRR $>$ 50\%. Overall, all categories exceed the 50\% threshold, showing that the erased model fails to suppress concepts embedded in latents inverted from a different backbone.
}
\label{tab:learned_embedding_different_backbone}
\end{table*}

\begin{table*}[h]
\centering
\resizebox{\textwidth}{!}{%
\begin{tabular}{llcrrrrrrrrrrr}
\toprule
Methods & Prompt & IRECE & \textit{airplane} & \textit{automobile} & \textit{bird} & \textit{cat} & \textit{deer} & \textit{dog} & \textit{frog} & \textit{horse} & \textit{ship} & \textit{truck} & \textbf{Avg.} \\
\midrule
\multirow{14.5}{*}{ESD~\cite{gandikota2023esd}} 
& \multirow{3.5}{*}{``''} & ~ & 88.67 & 98.00 & 89.33 & 94.67 & 94.67 & 98.00 & 92.67 & 95.33 & 86.67 & 90.67 & \textbf{92.87} \\
& ~ & \checkmark & 24.00 & 62.67 & 28.00 & 42.00 & 20.67 & 40.67 & 50.67 & 32.67 & 40.67 & 9.33 & \textbf{35.14} \\
\cmidrule(lr){3-14}
& ~ & $\Delta$ & \improve{-64.67} & \improve{-35.33} & \improve{-61.33} & \improve{-52.67} & \improve{-74.00} & \improve{-57.33} & \improve{-42.00} & \improve{-62.66} & \improve{-46.00} & \improve{-81.34} & \improve{-57.73} \\
\cmidrule(lr){2-14}
& \multirow{3.5}{*}{``image''} & ~ & 71.33 & 86.00 & 76.00 & 68.67 & 57.33 & 73.33 & 78.67 & 65.33 & 70.67 & 59.33 & \textbf{70.67} \\
& ~ & \checkmark & 27.33 & 62.67 & 20.00 & 33.33 & 19.33 & 31.33 & 39.33 & 30.67 & 36.67 & 10.67 & \textbf{31.13} \\
\cmidrule(lr){3-14}
& ~ & $\Delta$ & \improve{-44.00} & \improve{-23.33} & \improve{-56.00} & \improve{-35.34} & \improve{-38.00} & \improve{-42.00} & \improve{-39.34} & \improve{-34.66} & \improve{-34.00} & \improve{-48.66} & \improve{-39.53} \\
\cmidrule(lr){2-14}
& \multirow{3.5}{*}{``object''} & ~ & 72.67 & 80.67 & 52.00 & 48.67 & 38.67 & 39.33 & 68.00 & 48.00 & 60.00 & 48.67 & \textbf{55.67} \\
& ~ & \checkmark & 24.67 & 64.00 & 19.22 & 32.67 & 19.33 & 26.00 & 56.67 & 29.33 & 32.00 & 19.33 & \textbf{32.32} \\
\cmidrule(lr){3-14}
& ~ & $\Delta$ & \improve{-48.00} & \improve{-16.67} & \improve{-32.78} & \improve{-16.00} & \improve{-19.34} & \improve{-13.33} & \improve{-11.33} & \improve{-18.67} & \improve{-28.00} & \improve{-29.34} & \improve{-23.35} \\
\cmidrule(lr){2-14}
& \multirow{3.5}{*}{\texttt{TARGET}} & ~ & 68.00 & 86.67 & 52.00 & 50.67 & 35.33 & 58.67 & 58.67 & 44.00 & 63.33 & 46.00 & \textbf{56.33} \\
& ~ & \checkmark & 39.33 & 76.67 & 19.33 & 43.33 & 37.33 & 47.33 & 52.00 & 44.00 & 60.00 & 22.67 & \textbf{44.20} \\
\cmidrule(lr){3-14}
& ~ & $\Delta$ & \improve{-28.67} & \improve{-10.00} & \improve{-32.67} & \improve{-7.34} & \worse{+2.00} & \improve{-11.34} & \improve{-6.67} & 0.00 & \improve{-3.33} & \improve{-23.33} & \improve{-12.14} \\
\midrule
\multirow{14.5}{*}{UCE~\cite{gandikota2024uce}} 
& \multirow{3.5}{*}{``''} & ~ & 92.00 & 98.00 & 91.33 & 96.67 & 98.00 & 98.67 & 92.67 & 97.33 & 89.33 & 95.33 & \textbf{90.67} \\
& ~ & \checkmark & 37.33 & 82.00 & 48.67 & 71.33 & 44.00 & 70.67 & 64.67 & 46.00 & 53.33 & 26.67 & \textbf{54.47} \\
\cmidrule(lr){3-14}
& ~ & $\Delta$ & \improve{-54.67} & \improve{-16.00} & \improve{42.66} & \improve{-25.34} & \improve{-54.00} & \improve{-28.00} & \improve{-28.00} & \improve{-51.33} & \improve{-36.00} & \improve{-68.66} & \improve{-36.20} \\
\cmidrule(lr){2-14}
& \multirow{3.5}{*}{``image''} & ~ & 70.00 & 77.33 & 83.33 & 73.33 & 75.33 & 62.00 & 76.67 & 78.67 & 73.33 & 76.67 & \textbf{70.67} \\
& ~ & \checkmark & 28.00 & 49.33 & 26.00 & 51.33 & 29.33 & 31.33 & 58.00 & 38.67 & 38.67 & 22.67 & \textbf{37.33} \\
\cmidrule(lr){3-14}
& ~ & $\Delta$ & \improve{-42.00} & \improve{-28.00} & \improve{57.33} & \improve{-22.00} & \improve{-46.00} & \improve{-30.67} & \improve{-18.67} & \improve{-40.00} & \improve{-34.66} & \improve{-54.00} & \improve{-33.33} \\
\cmidrule(lr){2-14}
& \multirow{3.5}{*}{``object''} & ~ & 60.00 & 75.33 & 46.67 & 35.33 & 30.67 & 29.33 & 74.00 & 37.33 & 49.33 & 41.33 & \textbf{45.66} \\
& ~ & \checkmark & 28.67 & 71.33 & 18.00 & 22.00 & 17.33 & 17.33 & 56.00 & 32.00 & 38.00 & 19.33 & \textbf{32.00} \\
\cmidrule(lr){3-14}
& ~ & $\Delta$ & \improve{-31.33} & \improve{-4.00} & \improve{28.67} & \improve{-13.33} & \improve{-13.34} & \improve{-12.00} & \improve{-18.00} & \improve{-5.33} & \improve{-11.33} & \improve{-22.00} & \improve{-13.67} \\
\cmidrule(lr){2-14}
& \multirow{3.5}{*}{\texttt{TARGET}} & ~ & 84.67 & 96.67 & 90.67 & 94.67 & 86.67 & 94.67 & 91.33 & 92.67 & 88.00 & 92.00 & \textbf{86.87} \\
& ~ & \checkmark & 34.67 & 79.33 & 47.33 & 58.00 & 44.00 & 54.00 & 48.00 & 40.00 & 49.33 & 20.00 & \textbf{47.47} \\
\cmidrule(lr){3-14}
& ~ & $\Delta$ & \improve{-50.00} & \improve{-17.34} & \improve{43.34} & \improve{-36.67} & \improve{-42.67} & \improve{-40.67} & \improve{-43.33} & \improve{-52.67} & \improve{-38.67} & \improve{-72.00} & \improve{-39.40} \\
\midrule
\multirow{14.5}{*}{Receler~\cite{huang2024receler}} 
& \multirow{3.5}{*}{``''} & ~ & 91.33 & 100.00 & 90.67 & 94.67 & 97.33 & 98.67 & 92.00 & 96.67 & 88.00 & 94.67 & \textbf{94.40} \\
& ~ & \checkmark & 26.67 & 74.67 & 32.00 & 50.67 & 31.33 & 46.67 & 54.00 & 39.33 & 46.67 & 20.00 & \textbf{42.20} \\
\cmidrule(lr){3-14}
& ~ & $\Delta$ & \improve{-64.66} & \improve{-25.33} & \improve{-58.67} & \improve{-44.00} & \improve{-66.00} & \improve{-52.00} & \improve{-38.00} & \improve{-57.34} & \improve{-41.33} & \improve{-74.67} & \improve{-52.20} \\
\cmidrule(lr){2-14}
& \multirow{3.5}{*}{``image''} & ~ & 72.67 & 84.67 & 82.00 & 74.67 & 68.67 & 82.00 & 80.67 & 78.00 & 68.67 & 80.67 & \textbf{77.27} \\
& ~ & \checkmark & 28.00 & 56.00 & 34.00 & 42.00 & 22.00 & 40.67 & 48.00 & 39.33 & 32.00 & 18.00 & \textbf{36.00} \\
\cmidrule(lr){3-14}
& ~ & $\Delta$ & \improve{-44.67} & \improve{-28.67} & \improve{-48.00} & \improve{-32.67} & \improve{-46.67} & \improve{-41.33} & \improve{-32.67} & \improve{-38.67} & \improve{-36.67} & \improve{-62.67} & \improve{-41.27} \\
\cmidrule(lr){2-14}
& \multirow{3.5}{*}{``object''} & ~ & 67.33 & 72.67 & 54.00 & 50.00 & 48.00 & 38.67 & 77.33 & 42.67 & 63.33 & 54.67 & \textbf{56.87} \\
& ~ & \checkmark & 32.67 & 61.33 & 22.00 & 28.67 & 21.33 & 27.33 & 59.33 & 32.00 & 41.33 & 26.00 & \textbf{35.20} \\
\cmidrule(lr){3-14}
& ~ & $\Delta$ & \improve{-34.66} & \improve{-11.34} & \improve{-32.00} & \improve{-21.33} & \improve{-26.67} & \improve{-11.34} & \improve{-18.00} & \improve{-10.67} & \improve{-22.00} & \improve{-28.67} & \improve{-21.67} \\
\cmidrule(lr){2-14}
& \multirow{3.5}{*}{\texttt{TARGET}} & ~ & 61.33 & 91.33 & 60.67 & 70.00 & 26.67 & 62.00 & 91.33 & 92.67 & 88.00 & 56.00 & \textbf{70.00} \\
& ~ & \checkmark & 21.33 & 80.67 & 16.00 & 40.67 & 17.33 & 26.67 & 48.00 & 40.00 & 49.33 & 20.67 & \textbf{36.07} \\
\cmidrule(lr){3-14}
& ~ & $\Delta$ & \improve{-40.00} & \improve{-10.66} & \improve{-44.67} & \improve{-29.33} & \improve{-9.34} & \improve{-35.33} & \improve{-43.33} & \improve{-52.67} & \improve{-38.67} & \improve{-35.33} & \improve{-33.93} \\
\bottomrule
\end{tabular}
}
\caption{Effect of IRECE on per-class CRR (\%) under the \textbf{white-box} latent inversion setting. Each block reports results for concept erasure methods without IRECE, with IRECE, and the corresponding change $\Delta$ (with IRECE minus without IRECE). A more negative $\Delta$ indicates stronger suppression of the target concept.}
\label{tab:irece_white}
\end{table*}

\begin{table*}[h]
\centering
\resizebox{\textwidth}{!}{%
\begin{tabular}{llcrrrrrrrrrrr}
\toprule
Methods & Prompt & IRECE & \textit{airplane} & \textit{automobile} & \textit{bird} & \textit{cat} & \textit{deer} & \textit{dog} & \textit{frog} & \textit{horse} & \textit{ship} & \textit{truck} & \textbf{Avg.} \\
\midrule
\multirow{14.5}{*}{ESD~\cite{gandikota2023esd}} 
& \multirow{3.5}{*}{``''} & ~ & 71.33 & 81.33 & 56.00 & 48.67 & 34.00 & 58.00 & 81.33 & 58.00 & 54.67 & 30.00 & \textbf{57.33} \\
& ~ & \checkmark & 23.33 & 62.00 & 20.00 & 30.00 & 11.33 & 32.00 & 43.33 & 23.33 & 39.33 & 5.33 & \textbf{29.00} \\
\cmidrule(lr){3-14}
& ~ & $\Delta$ & \improve{-48.00} & \improve{-19.33} & \improve{-36.00} & \improve{-18.67} & \improve{-22.67} & \improve{-26.00} & \improve{-38.00} & \improve{-34.67} & \improve{-15.34} & \improve{-24.67} & \improve{-28.33} \\
\cmidrule(lr){2-14}
& \multirow{3.5}{*}{``image''} & ~ & 51.33 & 73.33 & 24.67 & 34.67 & 16.00 & 40.00 & 62.00 & 29.33 & 39.33 & 36.00 & \textbf{40.67} \\
& ~ & \checkmark & 20.67 & 51.33 & 14.00 & 20.00 & 14.00 & 18.00 & 36.67 & 22.00 & 30.67 & 6.67 & \textbf{23.40} \\
\cmidrule(lr){3-14}
& ~ & $\Delta$ & \improve{-30.66} & \improve{-22.00} & \improve{-10.67} & \improve{-14.67} & \improve{-2.00} & \improve{-22.00} & \improve{-25.33} & \improve{-7.33} & \improve{-8.66} & \improve{-29.33} & \improve{-17.27} \\
\cmidrule(lr){2-14}
& \multirow{3.5}{*}{``object''} & ~ & 60.67 & 83.33 & 37.33 & 33.33 & 32.00 & 46.00 & 80.00 & 43.33 & 58.67 & 56.67 & \textbf{53.13} \\
& ~ & \checkmark & 30.67 & 68.67 & 24.67 & 28.00 & 24.67 & 24.00 & 55.33 & 42.67 & 36.67 & 24.67 & \textbf{35.74} \\
\cmidrule(lr){3-14}
& ~ & $\Delta$ & \improve{-30.00} & \improve{-14.66} & \improve{-12.66} & \improve{-5.33} & \improve{-7.33} & \improve{-22.00} & \improve{-24.67} & \improve{-0.66} & \improve{-22.00} & \improve{-32.00} & \improve{-17.39} \\
\cmidrule(lr){2-14}
& \multirow{3.5}{*}{\texttt{TARGET}} & ~ & 46.67 & 74.00 & 16.67 & 32.00 & 35.33 & 34.00 & 57.33 & 27.33 & 32.67 & 38.67 & \textbf{39.47} \\
& ~ & \checkmark & 34.00 & 68.67 & 16.00 & 25.33 & 25.33 & 28.00 & 49.33 & 25.33 & 28.67 & 20.00 & \textbf{32.07} \\
\cmidrule(lr){3-14}
& ~ & $\Delta$ & \improve{-12.67} & \improve{-5.33} & \improve{-0.67} & \improve{-6.67} & \improve{-10.00} & \improve{-6.00} & \improve{-8.00} & \improve{-2.00} & \improve{-4.00} & \improve{-18.67} & \improve{-7.40} \\
\midrule
\multirow{14.5}{*}{UCE~\cite{gandikota2024uce}}
& \multirow{3.5}{*}{``''} & ~ & 91.33 & 98.00 & 92.67 & 96.67 & 98.00 & 98.67 & 93.33 & 97.33 & 90.00 & 96.00 & \textbf{95.20} \\
& ~ & \checkmark & 37.33 & 84.00 & 46.00 & 68.67 & 47.33 & 70.67 & 65.33 & 44.00 & 56.00 & 28.00 & \textbf{54.73} \\
\cmidrule(lr){3-14}
& ~ & $\Delta$ & \improve{-54.00} & \improve{-14.00} & \improve{-46.67} & \improve{-28.00} & \improve{-50.67} & \improve{-28.00} & \improve{-28.00} & \improve{-53.33} & \improve{-34.00} & \improve{-68.00} & \improve{-40.47} \\
\cmidrule(lr){2-14}
& \multirow{3.5}{*}{``image''} & ~ & 54.00 & 70.00 & 32.67 & 60.67 & 54.67 & 41.33 & 75.33 & 43.33 & 34.67 & 92.00 & \textbf{58.87} \\
& ~ & \checkmark & 16.67 & 45.33 & 23.33 & 40.00 & 21.33 & 26.00 & 51.33 & 35.33 & 31.33 & 34.00 & \textbf{33.40} \\
\cmidrule(lr){3-14}
& ~ & $\Delta$ & \improve{-37.33} & \improve{-24.67} & \improve{-9.34} & \improve{-20.67} & \improve{-33.34} & \improve{-15.33} & \improve{-24.00} & \improve{-8.00} & \improve{-3.34} & \improve{-58.00} & \improve{-25.47} \\
\cmidrule(lr){2-14}
& \multirow{3.5}{*}{``object''} & ~ & 31.33 & 74.00 & 20.67 & 8.00 & 15.33 & 21.33 & 63.33 & 30.00 & 37.33 & 68.00 & \textbf{36.93} \\
& ~ & \checkmark & 13.33 & 65.33 & 12.67 & 7.33 & 8.00 & 13.33 & 50.00 & 28.67 & 32.00 & 29.33 & \textbf{26.00} \\
\cmidrule(lr){3-14}
& ~ & $\Delta$ & \improve{-18.00} & \improve{-8.67} & \improve{-8.00} & \improve{-0.67} & \improve{-7.33} & \improve{-8.00} & \improve{-13.33} & \improve{-1.33} & \improve{-5.33} & \improve{-38.67} & \improve{-10.93} \\
\cmidrule(lr){2-14}
& \multirow{3.5}{*}{\texttt{TARGET}} & ~ & 40.67 & 74.00 & 11.33 & 28.00 & 30.00 & 30.00 & 50.00 & 25.33 & 24.67 & 33.33 & \textbf{34.73} \\
& ~ & \checkmark & 26.00 & 63.33 & 11.33 & 14.67 & 12.67 & 23.33 & 44.00 & 18.67 & 24.00 & 19.33 & \textbf{25.73} \\
\cmidrule(lr){3-14}
& ~ & $\Delta$ & \improve{-14.67} & \improve{-10.67} & 0.00 & \improve{-13.33} & \improve{-17.33} & \improve{-6.67} & \improve{-6.00} & \improve{-6.66} & \improve{-0.67} & \improve{-14.00} & \improve{-9.00} \\
\midrule
\multirow{14.5}{*}{Receler~\cite{huang2024receler}} 
& \multirow{3.5}{*}{``''} & ~ & 84.00 & 87.33 & 82.00 & 68.00 & 66.67 & 82.67 & 84.00 & 84.67 & 77.33 & 73.33 & \textbf{79.00} \\
& ~ & \checkmark & 24.00 & 62.67 & 39.33 & 43.33 & 27.33 & 43.33 & 44.00 & 40.00 & 42.00 & 13.33 & \textbf{37.93} \\
\cmidrule(lr){3-14}
& ~ & $\Delta$ & \improve{-60.00} & \improve{-24.66} & \improve{-42.67} & \improve{-24.67} & \improve{-39.34} & \improve{-39.34} & \improve{-40.00} & \improve{-44.67} & \improve{-35.33} & \improve{-60.00} & \improve{-41.07} \\
\cmidrule(lr){2-14}
& \multirow{3.5}{*}{``image''} & ~ & 43.33 & 73.33 & 36.67 & 40.00 & 26.00 & 50.00 & 56.67 & 43.33 & 27.33 & 51.33 & \textbf{44.80} \\
& ~ & \checkmark & 16.00 & 52.00 & 20.00 & 20.00 & 12.67 & 30.67 & 37.33 & 30.67 & 26.67 & 13.33 & \textbf{25.93} \\
\cmidrule(lr){3-14}
& ~ & $\Delta$ & \improve{-27.33} & \improve{-21.33} & \improve{-16.67} & \improve{-20.00} & \improve{-13.33} & \improve{-19.33} & \improve{-19.34} & \improve{-12.66} & \improve{-0.66} & \improve{-38.00} & \improve{-18.87} \\
\cmidrule(lr){2-14}
& \multirow{3.5}{*}{``object''} & ~ & 70.00 & 66.67 & 33.33 & 33.33 & 46.00 & 48.00 & 71.33 & 40.67 & 56.67 & 78.00 & \textbf{54.50} \\
& ~ & \checkmark & 36.67 & 60.00 & 23.33 & 26.67 & 30.00 & 32.00 & 57.33 & 30.00 & 32.00 & 32.67 & \textbf{36.07} \\
\cmidrule(lr){3-14}
& ~ & $\Delta$ & \improve{-33.33} & \improve{-6.67} & \improve{-10.00} & \improve{-6.66} & \improve{-16.00} & \improve{-16.00} & \improve{-14.00} & \improve{-10.67} & \improve{-24.67} & \improve{-45.33} & \improve{-18.43} \\
\cmidrule(lr){2-14}
& \multirow{3.5}{*}{\texttt{TARGET}} & ~ & 28.67 & 69.33 & 12.00 & 17.33 & 23.33 & 12.00 & 50.67 & 23.33 & 24.67 & 26.00 & \textbf{28.00} \\
& ~ & \checkmark & 14.67 & 60.00 & 7.33 & 12.67 & 11.33 & 5.33 & 53.33 & 21.33 & 24.00 & 12.00 & \textbf{21.07} \\
\cmidrule(lr){3-14}
& ~ & $\Delta$ & \improve{-14.00} & \improve{-9.33} & \improve{-4.67} & \improve{-4.66} & \improve{-12.00} & \improve{-6.67} & \worse{+2.66} & \improve{-2.00} & \improve{-0.67} & \improve{-14.00} & \improve{-6.93} \\
\bottomrule
\end{tabular}
}
\caption{Effect of IRECE on per-class CRR (\%) under the \textbf{black-box} latent inversion setting. Each block reports results for concept erasure methods without IRECE, with IRECE, and the corresponding change $\Delta$ (with IRECE minus without IRECE). A more negative $\Delta$ indicates stronger suppression of the target concept.}
\label{tab:irece_black}
\end{table*}

\end{document}